\documentclass[10pt,twocolumn,letterpaper]{article}
\usepackage[pagenumbers]{cvpr}
\usepackage{iftex}
\ifXeTeX
  \usepackage[no-math]{fontspec}
\else
  \usepackage[utf8]{inputenc}
  \usepackage[T1]{fontenc}
\fi
\usepackage{url}
\usepackage{booktabs}
\usepackage{amsfonts}
\usepackage{amsmath}
\usepackage{amssymb}
\usepackage{nicefrac}
\usepackage{microtype}
\usepackage{xcolor}
\usepackage[table]{colortbl}
\usepackage{multirow}
\usepackage{makecell}
\usepackage{graphicx}
\usepackage{wrapfig}
\usepackage{float}
\makeatletter
\providecommand{\FloatBarrier}{\par\begingroup\let\@elt\relax\edef\@tempa{\@botlist\@deferlist\@dbldeferlist}\ifx\@tempa\@empty\else\clearpage\fi\endgroup}
\makeatother
\usepackage{subcaption}
\usepackage{array}
\usepackage{enumitem}
\usepackage[most]{tcolorbox}
\usepackage{fvextra}

\def\name{MSBench}

\definecolor{rankA}{RGB}{72,  152, 82}
\definecolor{rankB}{RGB}{118, 190, 126}
\definecolor{rankD}{RGB}{208, 240, 212}
\definecolor{rankE}{RGB}{253, 192, 140}
\definecolor{rankF}{RGB}{254, 212, 170}
\definecolor{rankG}{RGB}{254, 232, 200}
\definecolor{rankH}{RGB}{255, 245, 230}
\definecolor{categorybg}{RGB}{240,240,240}
\newcommand{\rankbox}[1]{\textcolor{#1}{\rule{0.7em}{0.7em}}}
\definecolor{grpConv}{HTML}{D1495B}
\definecolor{grpNative}{HTML}{00798C}
\definecolor{grpSpec}{HTML}{5C4D7D}
\newcommand{\cell}[2]{\cellcolor{#1}#2}
\newcommand{\best}[2]{\cellcolor{#1}\textbf{#2}}



\newcommand{\numDDcv}{60.7}
\newcommand{\numDDcvModel}{69.4}
\newcommand{\numDDcvPrecise}{0.607}
\newcommand{\numDDcvModelPrecise}{0.694}
\newcommand{\numStabBase}{0.95}
\newcommand{\numStabFifteen}{0.94}

\newcommand{\numFinalBT}{0.87}
\newcommand{\numFinalBTk}{0.74}

\newcommand{\numPairN}{600}
\newcommand{\numPairPerSysLo}{67}
\newcommand{\numPairPerSysHi}{104}

\newcommand{\numJudgeModLo}{0.75}
\newcommand{\numJudgeModHi}{0.98}
\newcommand{\numICCLo}{0.68}
\newcommand{\numICCHi}{0.97}
\newcommand{\numJudgeFinal}{0.996}

\definecolor{cvprblue}{rgb}{0.21,0.49,0.74}
\usepackage[pagebackref,breaklinks,colorlinks,allcolors=cvprblue]{hyperref}
\def\confName{CVPR}
\def\confYear{2027}
\newcommand{\projecturl}{\url{https://totalhuang.github.io/MSBench-page/}}

\title{\name{}: Benchmarking Script-Controlled Multi-Shot Text-to-Video Generation}

\author{%
  Tao Huang$^{1}$, ZiHao Liang$^{1}$, Junhao Liu$^{2}$, Junzhe Li$^{2}$, Yuancheng Wei$^{2}$\\
  Sifan Zuo$^{1}$, Beiming Yu$^{1}$, Taojin Liu$^{1}$, Zhuoyang Li$^{1}$, Bing Cao$^{1}$\\[0.5em]
  $^{1}$Tianjin University \qquad
  $^{2}$Peking University
}

\begin{document}
\maketitle
\begin{abstract}
Text-to-video (T2V) models can now generate visually plausible short clips, but multi-shot generation requires a different set of skills: composing multiple shots with intended cut timing, transition style, camera language, and cross-shot continuity. How far this multi-shot output can be controlled from text remains unmeasured: existing benchmarks either evaluate single shots or, when they provide shot-level scripts, typically check whether the requested shots appear but not when they change or how they are joined. We introduce \textbf{\name{}}, a benchmark for controllable text-driven multi-shot video generation built around structured Scene-Entity-Shot mini-scripts that specify each shot's time span, camera state, content, and outgoing transition. \name{} defines a four-dimensional hybrid evaluation framework spanning Visual Quality, Dynamic Quality, Semantic Alignment, and Cross-Shot Consistency with 11 sub-metrics. Experiments on 14 commercial and open-source systems show that visual quality does little to separate the systems, whereas multi-shot control does: many models cut mainly on scene changes and render requested gradual transitions as hard cuts. Human evaluations further validate \name{}, with its system ranking reaching a Spearman correlation of \textbf{\numFinalBT{}} with human preferences. The project page is available at \projecturl{}.
\end{abstract}

\section{Introduction}
\label{sec:intro}

Recent text-to-video (T2V) models~\citep{sora2024,google_veo3_2025,kong2024hunyuanvideo,wu2025hunyuanvideo,wan2025wan,chen2025skyreels} produce convincing single-shot clips, but multi-shot generation is not simply a longer version of the same task. A multi-shot video additionally involves where shots change, how they are joined, which camera scale, angle, and movement each shot uses, and which entities and scenes reappear across them. Open-source multi-shot generators~\citep{meng2025holocine,wang2025echoshot,kara2025shotadapter,wu2025cinetrans,wang2025multishotmaster,zheng2024videogen} target this setting, and recent commercial systems such as Seedance~2.0~\citep{seedance2} and Kling~3.0~\citep{kling3} support multi-shot generation natively. With these systems, users can obtain complex multi-shot videos by writing a shot plan as a text prompt, without actual filming.

\begin{figure*}[!t]
  \centering
  \includegraphics[width=0.9\linewidth]{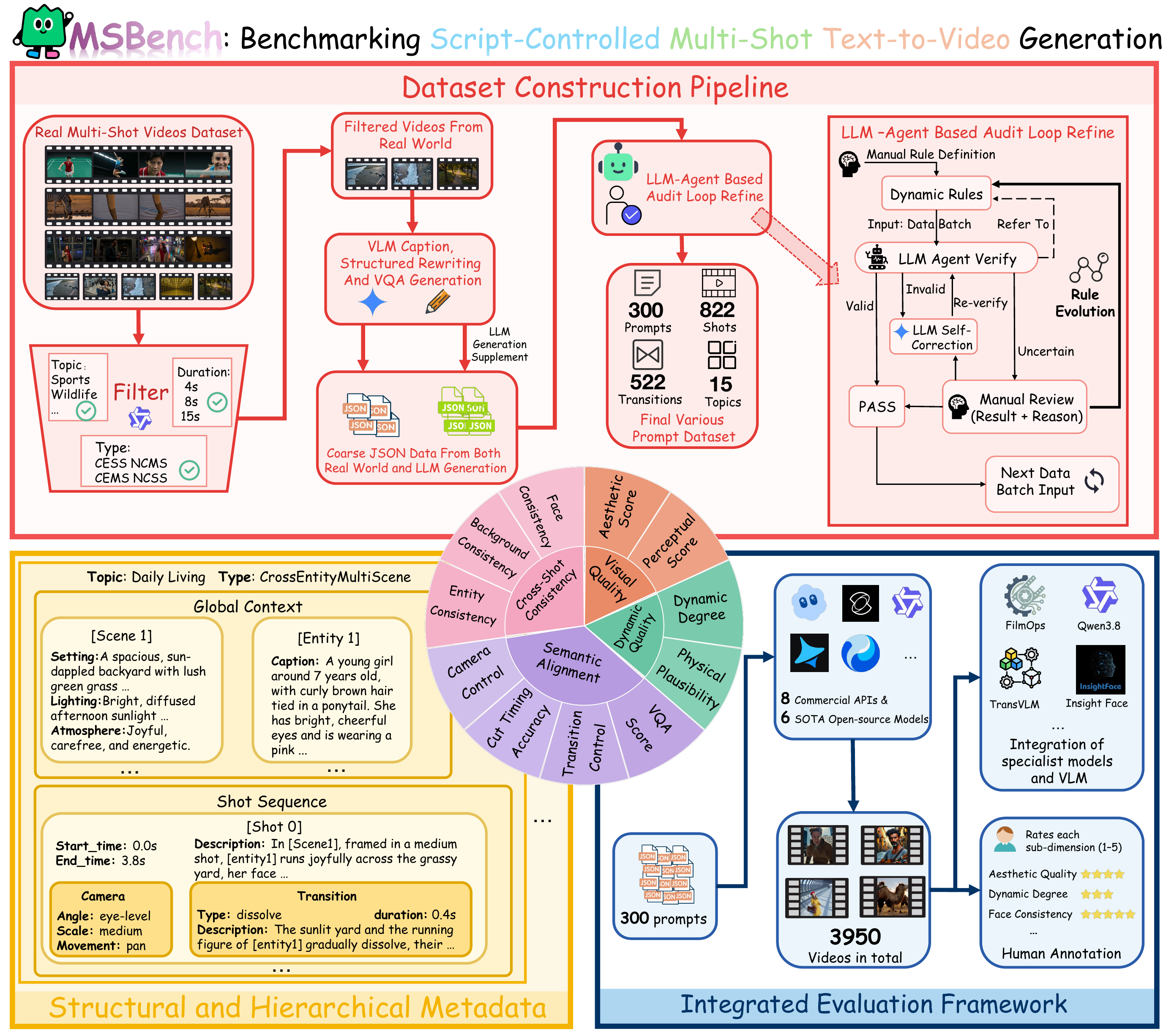}
  \caption{\textbf{Overview of \name{}.} MSBench is a benchmark for text-to-video multi-shot generation. It constructs hierarchical Scene-Entity-Shot metadata from real-world videos and LLM-generated supplements, then uses a hybrid evaluation framework to assess current T2V systems across visual quality, dynamic quality, semantic alignment, and cross-shot consistency.}
  \label{fig:overview}
\end{figure*}

A multi-shot prompt is essentially a shot-level script. Controllable multi-shot generation requires the system to follow this script in three aspects. The first is shot structure: the video should contain the requested number of shots, cut at the requested times, and use the requested transitions. The second is per-shot content: each shot should show the described content with the requested camera scale, angle, and movement. The third is cross-shot consistency: entities and scenes that appear in several shots should remain consistent. In applications such as short dramas and storyboard previsualization, a video with a missing cut, a wrong camera angle, or a character that changes appearance between shots cannot be used as generated. How well current systems follow a multi-shot script in these three aspects is a practical question for both model developers and users.

However, existing benchmarks cannot fully answer this question. General-purpose T2V benchmarks~\citep{huang2024vbench,liu2024evalcrafter,chen2026t2vworldbench,wang2025videoverse} assume each video is a single shot in both their prompts and metrics. Long-video benchmarks~\citep{zheng2025locot2v} extend the temporal horizon, but often treat a video as a continuous paragraph-level output without explicitly specified shots. Story-centric benchmarks~\citep{zhuang2025vistorybench,shi2026msvbench} introduce shot-level scripts, but often rely on character reference images, limiting their coverage of general text-driven T2V generation. Recent efforts further extend the evaluation to joint audio-video multi-shot generation~\citep{wei2026msavbench}. They measure how many of the requested shots are produced, but provide no explicit metrics for cut timing or transition type. Shot count alone cannot stand in for these metrics: we find that some systems achieve over 95\% correct shot counts yet still render more than 90\% of requested gradual transitions as hard cuts. A benchmark that evaluates all three aspects against the same script is still lacking.

To address this gap, we introduce \name{} (Figure~\ref{fig:overview}). Our contributions are as follows:
\begin{itemize}
    \item \textbf{A benchmark for controllable text-driven multi-shot video generation.} \name{} evaluates text-driven multi-shot video generation on shot structure, per-shot content, and cross-shot consistency against the same timestamped shot-level script. Its 300 Scene-Entity-Shot scripts make it possible to measure how well current multi-shot T2V systems can be controlled from text, down to the cut timing, transition type, and camera state of each shot.
    \item \textbf{A hybrid evaluation framework validated against human ratings.} We define 11 metrics across Visual Quality, Dynamic Quality, Semantic Alignment, and Cross-Shot Consistency, combining specialist perception models with powerful VLM judges. Notably, this includes multi-shot-specific metrics such as Cut Timing Accuracy and Transition Control. Extensive human studies confirm its reliability, with the overall system ranking reaching a Spearman correlation of \numFinalBT{} with human preferences.
    \item \textbf{A diagnostic study of current T2V systems.} We evaluate 14 commercial and open-source systems on 3{,}950 videos. Many general T2V systems cut less often when consecutive shots share a scene, rendering up to 47 points fewer of the requested cuts. Some systems with accurate shot counts still replace most requested gradual transitions with hard cuts. We further show that multiplying shot completion into each metric's score can mask opposite failure modes, and therefore report shot validity and the conditional score separately.
\end{itemize}

\section{Related Work}
\label{sec:related}

\paragraph{Multi-shot video generation.}
Early multi-shot systems follow a divide-and-conquer strategy. One line uses LLMs or agents to plan per-shot scripts and then generates and concatenates independent T2V clips~\citep{hu2024storyagent,wu2025automated,huang2025filmaster,shi2025animaker,zhuang2024vlogger,xie2024dreamfactory,lin2023videodirectorgpt,zhu2023moviefactory}; another generates anchor keyframes with T2I models and animates them with I2V models to keep identities consistent~\citep{zhao2024moviedreamer,long2024videostudio,zheng2024videogen}. End-to-end generators~\citep{meng2025holocine,wu2025cinetrans,wang2025echoshot,wang2025multishotmaster,kara2025shotadapter} instead produce the whole multi-shot video directly from a single prompt. Despite this progress, evaluation has lagged behind generation. Video-level metrics such as CLIP similarity cannot localize failures to individual shots or shot boundaries, and thus cannot tell whether a generated video follows the requested shot plan.

\begin{figure}[t]
  \centering
  \includegraphics[width=\linewidth]{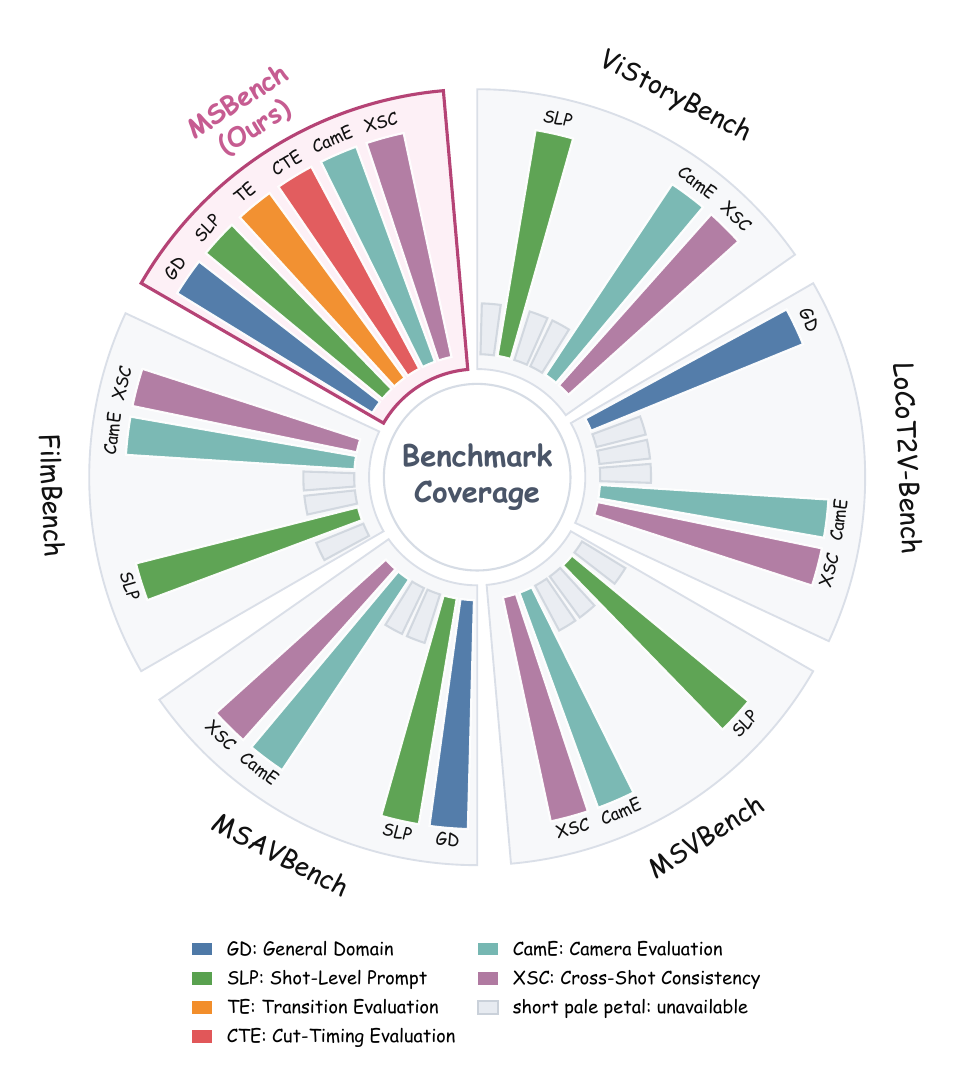}
  \caption{Comparison of representative benchmarks. Filled petals indicate supported capabilities: general domains, shot-level prompts, transition/cut-timing/camera evaluation, and cross-shot consistency.}
  \label{fig:bench_petal}
\end{figure}

\paragraph{Video-generation benchmarks.} Early video-generation benchmarks~\citep{huang2024vbench,liu2024evalcrafter} rely primarily on specialist perceptual models such as CLIP~\citep{radford2021learning} and DINO~\citep{caron2021emerging}. With recent advances in vision-language models (VLMs) and their growing reasoning capabilities~\citep{bai2025qwen3,qwen38}, newer benchmarks~\citep{chen2026t2vworldbench,zheng2025vbench} increasingly adopt VLMs as judges, prompting them either to rate a video along given criteria~\citep{han2025video,wang2025lovebenchmarkingevaluatingtexttovideo} or to answer binary questions derived from the prompt~\citep{lin2024evaluating,wang2025videoverse}. However, their test cases and metrics remain confined to single-shot clips. Long-video benchmarks such as LoCoT2V-Bench~\citep{zheng2025locot2v} incorporate multi-shot criteria, but lack shot-level prompts and cannot diagnose per-shot fidelity. Story-centric benchmarks~\citep{bugliarello2023storybench,zhuang2025vistorybench,shi2026msvbench} introduce multi-shot scripts, yet they typically emphasize character-driven narratives or depend on visual reference prompts, leaving general-domain text-driven multi-shot generation underexplored. More recently, MSAVBench~\citep{wei2026msavbench} extends evaluation to joint audio-video generation, and FilmBench~\citep{wang2026filmbench} assesses the quality of cinematic language on film-derived shot lists, yet neither provides explicit metrics for when shots change or which transition joins them. In contrast, \name{} establishes a unified protocol for evaluating general text-driven multi-shot T2V generation. Specifically, it assesses shot structure (shot count, cut timing, and transitions), per-shot content and camera control, and cross-shot consistency. Figure~\ref{fig:bench_petal} contrasts the evaluation scope of \name{} with prior benchmarks.

\section{The \name{} Dataset}
\label{sec:dataset}

In this section, we describe the hierarchical prompts used for multi-shot generation (\ref{sec:dataset_schema}) and the pipeline for scalable dataset construction (\ref{sec:dataset_construction}).

\subsection{Hierarchical Structured Prompts}
\label{sec:dataset_schema}
Evaluating controllable cuts and transitions needs prompts that state these explicitly. Each entry is thus represented as a JSON object structured into three coupled sets, $x=(\mathcal{C},\mathcal{E},\mathcal{S})$: \emph{Scenes}, \emph{Entities}, and \emph{Shots}. A scene $c_j\in\mathcal{C}$ defines environmental priors such as setting, lighting, and atmosphere. The entity table $\mathcal{E}=\{e_i\}_{i=1}^{N}$ indexes key objects and characters with textual descriptions. Scenes and entities are referenced via explicit IDs (\eg, \texttt{[Scene1]}, \texttt{[entity1]}) reused across shots. The shot sequence $\mathcal{S}=\{s_t\}_{t=1}^{T}$ specifies, for each shot, the camera configuration (angle, scale, and movement), action, timestamps, and transition to the next shot. Each entry also includes binary VQA questions, such as ``Is there a scene ...?'', whose constrained yes/no format mitigates free-form hallucination in VLMs. Each question is further annotated with the specific shots where its answer holds, enabling hierarchical, shot-level querying. By tying scenes, entities, and shots to explicit IDs, this schema reduces ambiguity and allows evaluation metrics to isolate failures to individual shots, transitions, or cross-shot relations (Section~\ref{sec:eval_sa}).

\subsection{Dataset Construction}
\label{sec:dataset_construction}

We define a topic and type grid to ensure controlled coverage across diverse scenarios. The topics, adapted from YouTube categories and refined for cinematic representativeness, comprise 15 sub-topics categorized into Human \& Social, Nature \& Biological, Urban \& Tech, and Creative \& Artistic. To grade the difficulty of cross-shot control, the type field crosses two orthogonal axes: whether an entity must persist across shots (cross-entity, CE, vs.\ no cross-entity, NC) and whether the shots span one or several scenes (single-scene, SS, vs.\ multi-scene, MS), yielding four levels (CESS, CEMS, NCSS, NCMS). The distributions of scripts over topics, types, and duration tiers, together with camera and transition statistics, are provided in Appendix~\ref{app:dataset_details}.

We filter candidates from a private real-world video database using Qwen3-VL~\citep{bai2025qwen3} to verify alignment with our predefined topics, types, and duration constraints (4\,s or 8\,s). For the selected videos, Gemini-3-Pro generates shot-level captions, which are then structured into the Scene-Entity-Shot JSON schema alongside corresponding VQA questions. This conversion process focuses on translating the physical structure and camera language of real footage into controllable, text-based prompts without introducing external content. This collection is further augmented with synthetic samples to balance the distribution of topics, types, and camera movements, creating a hybrid initial pool.

Finally, we refine this pool through an LLM-agent-based audit loop built on Gemini-3-Pro. Human authors establish a rule charter as a dynamic knowledge base for quality requirements. The agent evaluates each entry against this charter, automatically correcting clear violations and flagging ambiguous cases for manual review. These reviewed cases are then used to update the agent's knowledge, allowing it to refine the charter and expand its checks in subsequent rounds. As the charter matures, the need for manual intervention decreases, enabling efficient scaling. To further examine model performance on longer videos with more shots, we scale up the dataset by reusing this pipeline and its audit loop to construct 50 additional scripts for 15\,s videos. A final human inspection ensures that all prompts, metadata, and VQA questions are coherent, evaluable, and consistent.

The resulting dataset consists of 300 structured prompts encompassing 822 shots and 522 transitions with comprehensive statistics reported in Appendix~\ref{app:dataset_details}.

\section{The \name{} Evaluation Framework}
\label{sec:eval}

\begin{figure*}[!t]
  \centering
  \includegraphics[width=1.0\linewidth]{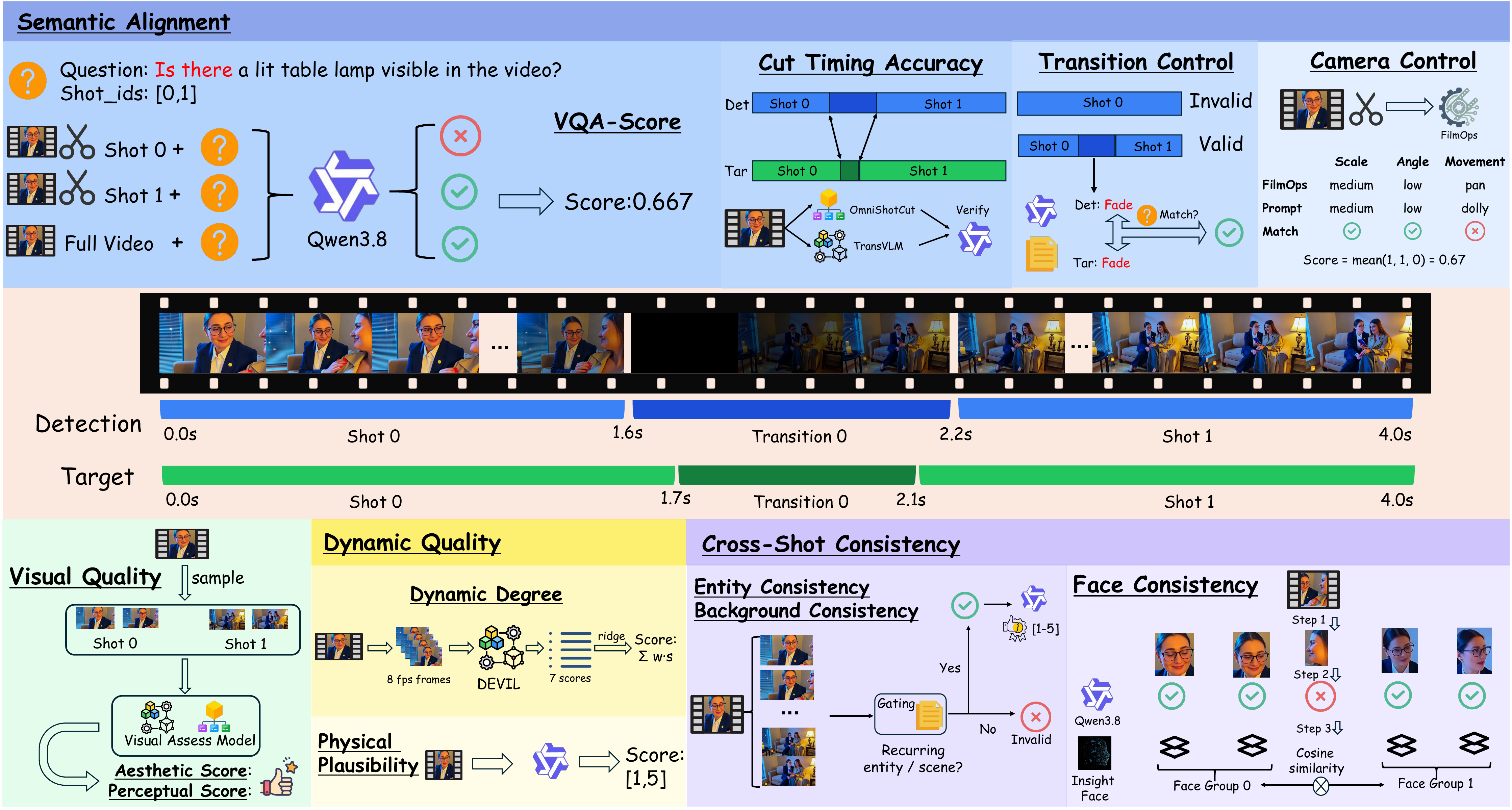}
  \caption{The \name{} evaluation framework.}
  \label{fig:pipeline}
\end{figure*}

\name{} evaluates a generated video along four dimensions (Figure~\ref{fig:pipeline}): \textbf{Visual Quality}, \textbf{Dynamic Quality}, \textbf{Semantic Alignment}, and \textbf{Cross-Shot Consistency}. Of the three aspects in Section~\ref{sec:intro}, Semantic Alignment covers shot structure (Transition Control, Cut Timing Accuracy) and per-shot content (VQA-Score, Camera Control), and Cross-Shot Consistency covers the third; Visual and Dynamic Quality are independent of the script. The framework pairs specialist perception models with VLM reasoning (Qwen3.8-27B~\citep{qwen38}), and a unified transition-recognition pipeline extracts the shot structure, decoupling within-shot quality from cross-shot control. Appendix~\ref{app:metric_details} details every sub-metric and its underlying model (Table~\ref{tab:metric_overview}), and Appendix~\ref{app:metric_vis} shows example videos for each sub-metric; the VLM prompt templates are released with the evaluation code.

For multi-shot metrics involving shot transitions, camera control, and cross-shot consistency, we report a validity rate (suffix \mbox{-V}), the share of the requested structure that is present and can be scored, alongside the quality score. This distinguishes whether performance deficits stem from a structural failure to produce multi-shot sequences or from deficient cinematic execution when such structures are present.

\subsection{Visual Quality}
\label{sec:eval_vq}
Visual Quality assesses aesthetic appeal and technical fidelity. \textbf{Aesthetic Score (AS)} evaluates composition, lighting, and color distribution via Q-Align~\citep{wu2023qalign}, while \textbf{Perceptual Score (PS)} detects technical artifacts and blur degradation via DeQA-Score~\citep{deqa_score}. To account for the multi-shot structure and avoid boundary transition artifacts, both metrics are computed at the individual shot level and subsequently aggregated across all shots.

\subsection{Dynamic Quality}
\label{sec:eval_dq}
Dynamic Quality asks whether the video contains meaningful motion and whether that motion is physically credible. \textbf{Dynamic Degree (DD)} measures subject and camera dynamics via DEVIL~\citep{liao2024devil}, calibrated against human motion ratings. Evaluating dynamics strictly within shot intervals prevents cuts and transitions from being falsely registered as motion. \textbf{Physical Plausibility (PP)} employs Qwen3.8~\citep{qwen38} to detect physical anomalies, including unnatural anatomical deformation, sudden object disappearance, and implausible interactions, complementing optical motion descriptors.

\subsection{Semantic Alignment}
\label{sec:eval_sa}
Semantic Alignment quantifies how faithfully the generated video realizes the structured script across event completeness, transition style, cut timing, and camera language.

\paragraph{VQA-Score (VQA).}
Binary question answering probes whether scripted events occur in the video~\citep{lin2024evaluating,wang2025videoverse}. To verify narrative chronology, \name{} associates each question with its target shot. The VLM evaluates both the designated shot intervals and the full video, awarding full credit only when events appear in their intended shots, thus penalizing temporal misplacement that global video VQA overlooks.

\paragraph{Transition Control (TC).}
TC assesses both the occurrence and type correctness of requested transitions. Because conventional detectors like PySceneDetect~\citep{castellano_pyscenedetect} and TransNet~V2~\citep{soucek2024transnet} used in prior work~\citep{meng2025holocine,kara2025shotadapter} struggle with gradual transitions, we combine OmniShotCut~\citep{wang2026omnishotcut} and TransVLM~\citep{chen2026transvlm}, with Qwen3.8 resolving ambiguous cases, to detect hard cuts, dissolves, fades, and wipes. These detected boundaries also define the shot intervals across our framework (see Appendix~\ref{app:transition_validation} for fusion rules and validation). Each requested transition is matched to a detected transition in temporal order. We report validity TC-V and type accuracy TC-A, with their product representing the overall transition score in Semantic Alignment.

\paragraph{Cut Timing Accuracy (CTA).}
CTA measures how closely the generated shot and transition boundaries follow the requested timeline. It operates in seconds rather than in frames as in earlier shot-timing metrics~\citep{meng2025holocine,kara2025shotadapter}, avoiding dependence on model-specific frame rates, and penalizes extra or missing shots as well as mistimed boundaries.

\paragraph{Camera Control (CC).}
CC evaluates whether each shot adheres to the requested camera angle, scale, and movement. Using detected shot boundaries, the framework segments the video and applies FilmOps cinematography operators~\citep{wang2026filmbench} to each shot individually, predicting shot scale and angle on sampled frames and camera movement from the clip. These attributes are matched against the script specifications to compute the camera score, with missing shots counted as zero and CC-V giving the share of scorable shots.

\subsection{Cross-Shot Consistency}
\label{sec:eval_csc}
Cross-Shot Consistency targets a defining challenge of multi-shot generation: preserving entities, backgrounds, and faces across cuts. Each sub-metric is computed only on prompts annotated with the corresponding consistency relation during dataset construction. Following our decoupled reporting, the validity rates (CSEC-V, CSBC-V, CSFC-V) give the share of videos that can be evaluated, which requires at least two detected shots, and the conditional scores (-S) are computed over valid videos.

\paragraph{Entity Consistency (CSEC).}
Qwen3.8 is given the predefined entity descriptions to locate recurring entities across shots, and rates on a 1--5 scale whether their shape, proportions, color, materials, and clothing remain consistent. Changes explained by viewpoint, pose, or the script are not counted as drift, and facial identity is left to CSFC.

\paragraph{Background Consistency (CSBC).}
Qwen3.8 compares only shots that the script assigns to the same scene, and rates on a 1--5 scale the continuity of spatial layout, fixed structures, and textures, without penalizing differences in viewpoint or lighting.

\paragraph{Face Consistency (CSFC).}
Faces require a dedicated pipeline. InsightFace~\citep{deng2019arcface} detects faces in each detected shot, Qwen3.8 selects the target person among the numbered face boxes, and the selected faces' identity features are compared across shots with calibrated nearest-neighbor similarity. Appendix~\ref{app:csfc} details the selection and scoring rules.

\section{Experiments}
\label{sec:experiments}

\begin{table*}[!t]
  \centering
  \caption{Results on the 250-script base set (4\,s and 8\,s); 0--100 scale, higher is better. Rank colors mark \rankbox{rankA} 1st, \rankbox{rankB} 2nd, and \rankbox{rankD} 3rd. Abbreviations: AS/PS, aesthetic/perceptual score; DD/PP, dynamic degree/physical plausibility; TC-V/TC-A, transition-control validity/accuracy; CTA, cut timing accuracy; CC-V/CC, camera-control validity/score; CSEC/CSBC/CSFC, cross-shot entity/background/face consistency (-V: validity, -S: conditional score on valid videos). $^\dagger$~Open-weight system.}
  \label{tab:main_results}
  \scriptsize
  \setlength{\tabcolsep}{3pt}
  \resizebox{\textwidth}{!}{%
  \begin{tabular}{l*{16}{c}c}
    \toprule
    \multirow{2}{*}{\textbf{Model}} & \multicolumn{2}{c}{\textbf{VQ}} & \multicolumn{2}{c}{\textbf{DQ}} & \multicolumn{6}{c}{\textbf{SA}} & \multicolumn{6}{c}{\textbf{CSC}} & \multirow{2}{*}{\textbf{Final}} \\
    \cmidrule(lr){2-3} \cmidrule(lr){4-5} \cmidrule(lr){6-11} \cmidrule(lr){12-17}
     & AS & PS & DD & PP & VQA & TC-V & TC-A & CTA & CC-V & CC & CSEC-V & CSEC-S & CSBC-V & CSBC-S & CSFC-V & CSFC-S & \\
    \midrule
    \multicolumn{18}{c}{\cellcolor{categorybg}\textbf{Conventional T2V Models}} \\
    Kling~2.6              & 95.7 & 75.8 & 65.3 & 86.3 & 69.8 & 34.7 & 35.0 & 31.7 & 64.1 & 30.0 & 30.5 & 81.9 & 21.9 & 82.7 & 17.3 & \cell{rankD}{66.0} & 53.9 \\
    Hailuo~2.3             & 88.7 & 67.0 & 66.9 & 87.7 & 65.0 & 20.7 & 55.9 & 24.3 & 56.1 & 26.8 & 16.8 & 92.5 & 12.3 & 84.9 & 16.0 & 65.4 & 49.8 \\
    Wan2.2$^\dagger$       & 94.4 & 76.7 & \cell{rankD}{68.8} & \cell{rankB}{88.5} & 70.5 & 47.3 & 37.3 & 35.1 & 71.3 & 31.4 & 43.5 & 92.5 & 38.1 & 75.7 & 32.1 & \cell{rankB}{74.5} & 58.5 \\
    HYVideo~1.5$^\dagger$  & 90.2 & 73.3 & 68.2 & 82.9 & 67.8 & 34.7 & 27.9 & 30.6 & 63.7 & 29.8 & 29.8 & 86.2 & 25.8 & 73.9 & 23.5 & 63.5 & 52.9 \\
    \midrule
    \multicolumn{18}{c}{\cellcolor{categorybg}\textbf{Native Multi-shot T2V Models}} \\
    Sora2~Pro              & 79.8 & 63.8 & 52.0 & 84.8 & 83.0 & \cell{rankB}{98.2} & 73.7 & \cell{rankD}{77.6} & \cell{rankB}{98.8} & 45.9 & \cell{rankB}{99.2} & 91.3 & \cell{rankD}{98.1} & 82.5 & 87.7 & 57.9 & 71.0 \\
    Veo3.1                 & \cell{rankD}{96.5} & \cell{rankD}{80.5} & 54.1 & 81.3 & 84.8 & 87.3 & \cell{rankB}{83.0} & 51.0 & 91.5 & 39.5 & 87.0 & 91.3 & 82.6 & \best{rankA}{88.7} & 79.0 & 57.9 & 71.1 \\
    Wan2.6                 & 96.3 & 79.1 & \best{rankA}{70.8} & 83.7 & 81.9 & 85.1 & 51.7 & 61.8 & 91.8 & 43.6 & 84.0 & \best{rankA}{94.9} & 78.7 & 83.2 & 66.7 & 59.9 & 71.1 \\
    Jimeng3.0~Pro          & \best{rankA}{97.5} & \best{rankA}{81.4} & 66.2 & 87.8 & 71.1 & 78.8 & 47.0 & 54.6 & 87.6 & 45.6 & 74.8 & 86.7 & 79.4 & 78.0 & 72.8 & 61.6 & 68.9 \\
    Seedance~2.0           & 93.5 & 74.3 & 61.7 & \best{rankA}{92.2} & \best{rankA}{88.6} & 97.1 & \cell{rankD}{80.2} & 70.0 & 98.4 & \best{rankA}{58.1} & \best{rankA}{100.0} & \cell{rankB}{94.1} & \cell{rankD}{98.1} & \cell{rankD}{86.1} & \cell{rankD}{93.8} & \best{rankA}{78.3} & \best{rankA}{79.6} \\
    Kling~3.0              & 89.3 & 72.2 & 65.8 & 88.0 & \cell{rankB}{87.1} & 97.4 & 39.7 & \cell{rankB}{82.4} & \cell{rankD}{98.5} & \cell{rankB}{57.8} & 97.7 & 92.0 & 97.4 & 85.8 & 77.8 & 58.6 & \cell{rankD}{74.3} \\
    MiniMax~H3$^\dagger$   & 94.1 & 77.3 & \cell{rankB}{69.8} & \cell{rankD}{88.2} & \cell{rankD}{86.6} & 95.2 & \best{rankA}{90.0} & 59.8 & 97.0 & \cell{rankD}{56.2} & 97.7 & \cell{rankD}{93.1} & 93.5 & \cell{rankB}{88.6} & \best{rankA}{96.3} & 65.3 & \cell{rankB}{78.9} \\
    \midrule
    \multicolumn{18}{c}{\cellcolor{categorybg}\textbf{Multi-shot Specialists}} \\
    HoloCine$^\dagger$     & 80.1 & 63.0 & 57.5 & 87.7 & 82.4 & 93.6 & 43.5 & 74.9 & 96.6 & 45.8 & 95.4 & 90.8 & 92.9 & 73.8 & 79.0 & 54.2 & 67.8 \\
    MSM$^\dagger$          & 90.2 & 70.8 & 43.3 & 81.6 & 80.6 & \best{rankA}{99.4} & 38.5 & \best{rankA}{87.0} & \best{rankA}{99.5} & 44.4 & \cell{rankB}{99.2} & 86.1 & \best{rankA}{99.4} & 73.3 & 66.7 & 40.5 & 66.8 \\
    VGoT$^\dagger$         & \cell{rankB}{96.8} & \cell{rankB}{81.2} & 45.3 & 63.1 & 69.7 & \cell{rankD}{97.8} & 27.9 & 69.3 & 98.3 & 42.1 & \cell{rankB}{99.2} & 47.4 & \best{rankA}{99.4} & 41.6 & \best{rankA}{96.3} & 48.1 & 60.0 \\
 
    \bottomrule
  \end{tabular}
  }
\end{table*}

\subsection{Setup}
\label{sec:setup}

We evaluate 14 T2V systems in three groups. \emph{Conventional T2V models} are relatively early general-purpose T2V models: Kling~2.6, Hailuo~2.3, Wan2.2~\citep{wan2025wan}, and HunyuanVideo-1.5~\citep{wu2025hunyuanvideo} (HYVideo~1.5). \emph{Native multi-shot T2V models} are general-purpose models that natively generate multiple shots from one prompt: Sora2~Pro, Veo3.1, Wan2.6, Jimeng3.0~Pro, Seedance~2.0~\citep{seedance2}, Kling~3.0~\citep{kling3}, and MiniMax~H3~\citep{minimax_h3}. \emph{Multi-shot specialists} are research systems that take per-shot inputs: the end-to-end generators HoloCine~\citep{meng2025holocine} and MultiShotMaster~\citep{wang2025multishotmaster} (MSM), and the agent-based pipeline VideoGen-of-Thought~\citep{zheng2024videogen} (VGoT). Eight systems are commercial and accessed through their official APIs; Wan2.2, HYVideo~1.5, MiniMax~H3, HoloCine, MSM, and VGoT are open-weight. All systems are evaluated on the 250-script base set, while the 50-script 15\,s extension is evaluated only on systems that natively support 15\,s generation, yielding 3{,}950 videos in total; results for the two parts are reported separately. Sub-metrics are reported on a 0--100 scale, and the Final Score is the equal-weight mean of the four dimension scores (Appendix~\ref{app:aggregation}).

\subsection{Main Results}
\label{sec:main_results}

Table~\ref{tab:main_results} reports the base set and Table~\ref{tab:results_15s} the 15\,s extension. Seedance~2.0 (79.6) and MiniMax~H3 (78.9) lead the base set, followed by Kling~3.0 (74.3). These are the three most recently released systems we evaluate, and their lead comes mainly from Semantic Alignment and Cross-Shot Consistency, while their visual quality is comparable to that of the other commercial systems. The open-weight MiniMax~H3 performs on par with the best commercial system. Wan2.6, Veo3.1, and Sora2~Pro form a close second group. The multi-shot specialists HoloCine and MSM receive per-shot lengths as input, which gives them high scores on structural metrics such as transition validity and cut timing, but they lag in motion, and HoloCine also in visual quality. The conventional T2V models often render a multi-shot script as a single shot, which lowers every structure-dependent metric. VGoT produces high-quality individual shots but the weakest cross-shot entity and background consistency, since its shots are generated independently. On the 15\,s extension, the leading systems remain close, while the two open-weight conventional T2V models fall further behind.

Overall, visual quality does little to separate the systems. The system-level rank correlation between Aesthetic Score and the Final Score is only 0.08, in line with earlier reports that aesthetics is a poor proxy for multi-shot control~\citep{shi2026msvbench}. For two systems on the same script, Semantic Alignment, which contains the structural sub-metrics, differs by 23.0 points on average, nearly three times the 7.9-point gap in Aesthetic Score. Current systems differ mainly in how they execute the script.

\begin{table}[!t]
  \centering
  \caption{Results on the 50-script 15\,s extension for the nine systems that natively generate 15\,s. Dimension and Final scores, 0--100; $^\dagger$~open-weight system.}
  \label{tab:results_15s}
  \footnotesize
  \setlength{\tabcolsep}{4pt}
  \begin{tabular}{@{\hspace{8pt}}w{l}{62pt}@{\hspace{14pt}}*{5}{w{c}{\dimexpr(\columnwidth-128pt)/5\relax}}@{\hspace{4pt}}}
    \toprule
    \textbf{Model} & \textbf{VQ} & \textbf{DQ} & \textbf{SA} & \textbf{CSC} & \textbf{Final} \\
    \midrule
    Wan2.2$^\dagger$       & 80.6 & 73.9 & 16.0 & 23.4 & 48.5 \\
    HYVideo~1.5$^\dagger$  & 84.5 & 62.5 & 15.8 & 21.3 & 46.0 \\
    Wan2.6                 & \cell{rankB}{87.2} & \cell{rankB}{77.8} & 61.7 & 81.5 & 77.1 \\
    Seedance~2.0           & 84.1 & 76.4 & \cell{rankB}{65.5} & \best{rankA}{90.6} & \cell{rankB}{79.2} \\
    Kling~3.0              & 79.3 & \cell{rankD}{77.6} & \best{rankA}{67.7} & \cell{rankD}{85.1} & \cell{rankD}{77.4} \\
    MiniMax~H3$^\dagger$   & \cell{rankD}{86.0} & \best{rankA}{81.1} & 62.7 & \cell{rankB}{88.3} & \best{rankA}{79.5} \\
    HoloCine$^\dagger$     & 69.4 & 69.3 & 61.8 & 74.0 & 68.6 \\
    MSM$^\dagger$          & 82.2 & 63.1 & \cell{rankD}{65.4} & 72.4 & 70.8 \\
    VGoT$^\dagger$         & \best{rankA}{89.7} & 51.6 & 55.3 & 43.4 & 60.0 \\
 
    \bottomrule
  \end{tabular}
\end{table}

\subsection{Findings}
\label{sec:insights}

\begin{figure*}[!t]
  \centering
  \includegraphics[width=\linewidth]{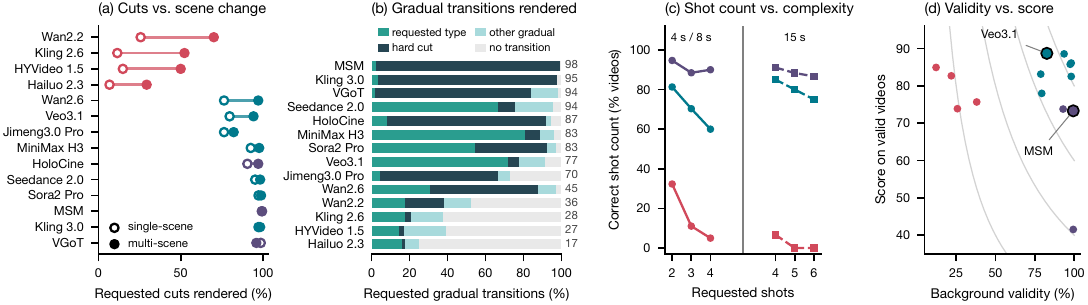}
  \caption{Evidence for the findings. (a)~Share of requested cuts that are rendered, in single-scene and multi-scene base-set scripts (Finding~1). (b)~Outcome of each requested gradual transition on the base set, per system; systems are sorted by shot-count accuracy, given on the right (Finding~2). (c)~Share of videos with the requested shot count, by number of requested shots, on the base set and the 15\,s extension (Finding~4). (d)~Background consistency on the base set: validity against the score on valid videos; gray curves connect equal products (Finding~5). Colors mark the system groups of Section~\ref{sec:setup}: \rankbox{grpConv}~conventional T2V, \rankbox{grpNative}~native multi-shot T2V, and \rankbox{grpSpec}~multi-shot specialists.}
  \label{fig:findings}
\end{figure*}

Beyond the overall ranking, we examine how the systems execute the shot-level script: where they place cuts, how they render transitions and camera states, how their structure holds up on more complex scripts, and how missing shots affect the scores. We summarize the results in five findings (Figure~\ref{fig:findings}). All intervals are 95\% bootstrap intervals over scripts, and the detailed numbers are given in Appendix~\ref{app:additional_results}.

\paragraph{Finding~1: Systems with weaker multi-shot control tend to cut at scene changes.}
We compare the share of requested cuts that are rendered in single-scene scripts, where consecutive shots keep the location, with multi-scene scripts (Figure~\ref{fig:findings}a). With duration, shot count, and transition type held fixed, six systems render 13 to 47 points fewer cuts in single-scene scripts, with the largest gaps for the four conventional T2V models, which also have the lowest shot-count accuracy, while the remaining eight, including Kling~3.0, Seedance~2.0, and the multi-shot specialists, show no significant gap (Appendix~\ref{app:scene_cut}). As an intervention, we rewrite 24 single-scene scripts so that the second shot moves to a new location (Appendix~\ref{app:c1_intervention}). Across Wan2.6, Kling~2.6, and Wan2.2, the new location adds a shot boundary in 32 pairs and removes one in 3 ($p<10^{-6}$), though most added boundaries are not the requested hard cut. Pooled over these systems, a change of location prompts a shot boundary, while the remaining systems show that cutting within a scene is learnable. Future multi-shot models therefore need to decouple camera switches from changes of environment, which calls for training on multi-angle coverage within shared scenes.

\paragraph{Finding~2: Accurate shot structure does not imply accurate transitions.}
Requested gradual transitions are most often rendered as hard cuts: a requested dissolve becomes a hard cut in 45.9\% of cases and a dissolve in only 28.5\% (Appendix~\ref{app:transition_breakdown}). This substitution does not shrink as shot structure improves (Figure~\ref{fig:findings}b). MSM and Kling~3.0 produce the requested shot count in 98\% and 95\% of videos, yet render more than 90\% of requested gradual transitions as hard cuts; their TC-A (38.5 and 39.7) is close to the 37.5 that a system rendering every transition as a hard cut would obtain. From Kling~2.6 to Kling~3.0, shot-count accuracy rises by 67 points while the share of gradual transitions rendered with the requested type falls from 17.5\% to 3.6\%. Learning where shots change does not teach a model how they should change. Transition type therefore needs its own conditioning and supervision, \eg, training clips with labeled transitions, beyond control over where shots change.

\paragraph{Finding~3: Camera control is reliable mainly for common camera states.}
Camera Control is low even for systems that render almost every requested shot, so the errors arise within the shots. Common states are followed far more often than rare ones: on shots that are rendered, eye-level reaches 76.0 and a static camera 73.6, whereas a tilted angle reaches 14.3 and a tracking shot 26.6. When the request is not met, the detected state is most often a static camera, a wider framing, or an eye-level view (Appendix~\ref{app:camera_breakdown}). Stronger per-shot camera control may therefore require rebalancing this long tail, for example through targeted data and camera annotations for rare viewpoints, scales, and movements, rather than only scaling generic video corpora.

\paragraph{Finding~4: Shot structure degrades on longer scripts with more shots.}
For conventional T2V models, the share of videos with the requested shot count falls from 32.3\% for two shots to 5.0\% for four, and on the 15\,s extension, where only the two open models are evaluated, 60--73\% of their videos are a single take (Figure~\ref{fig:findings}c). Native multi-shot T2V models hold up better but still decline from 81.3\% to 60.0\%, increasingly by adding shots, while the specialists stay near 90\%. Aesthetic Score is flat across all of these conditions (Appendix~\ref{app:shot_count}). This matches how conventional T2V models are trained: their data pipelines split raw videos into single-shot clips with a shot-boundary detector~\citep{kong2024hunyuanvideo,yang2025cogvideox,opensora2025v2}, and HYVideo~1.5 additionally removes clips that contain transitions~\citep{wu2025hunyuanvideo}, so cuts are largely absent from what they learn. Complex multi-shot capability therefore has to be acquired through dedicated training, and professional multi-shot generation may benefit from shot-level inputs that make the temporal layout explicit, rather than leaving it to be inferred from a single long prompt.

\paragraph{Finding~5: A combined score can mask opposite failure modes.}
Existing multi-shot benchmarks report one number per metric. MSAVBench, for example, scores the shots that remain after alignment and multiplies the overall score by the shot-completion rate~\citep{wei2026msavbench}, but does not report how many videos a consistency metric could score. Such a number can hide opposite failure modes (Figure~\ref{fig:findings}d). Multiplying our validity and conditional score, Veo3.1 and MSM obtain 73.3 and 72.9 on background consistency, yet Veo3.1 produces comparable shots in 82.6\% of scripts and keeps the background well (88.7), whereas MSM produces them in 99.4\% but preserves the background less (73.3). The two call for different remedies. Low-validity systems first need to realize the requested shots, whereas low-score systems need to carry identity and background across shots, \eg, through shared memory or reference conditioning. Similar ties occur for other metrics and system pairs (Appendix~\ref{app:validity_split}). We therefore encourage multi-shot benchmarks to report validity and the score on valid videos separately.

\subsection{Human Alignment and Judge Robustness}
\label{sec:human}

\paragraph{Protocol.}
We collect two kinds of human judgments on anonymized videos from all 14 systems, including 15\,s videos (Appendix~\ref{app:human_study}). First, sub-metric ratings cover 780 distinct videos (19.7\% of the 3{,}950 evaluated videos; 218 per dimension), which five annotators rate following the same rubrics as the automatic metrics. Second, three annotators judge which of two videos generated from the same script is better overall, on 540 pairs involving 1{,}025 videos (25.9\%). Together, 40.2\% of the evaluated videos receive human judgments. All annotators hold at least a bachelor's degree, are researchers working in computer vision, and receive a short unified training before annotating. The mean ratings are reliable, with an inter-annotator ICC of \numICCLo{}--\numICCHi{} across items (Appendix~\ref{app:human_correlation}).

\begin{table}[!t]
  \centering
  \scriptsize
  \setlength{\tabcolsep}{4pt}
  \caption{Agreement ($\times$100) with human ratings. SRCC, KRCC, and PLCC are video-level; \emph{Model} is the system-level SRCC. $^\dagger$Cross-validated by script.}
  \label{tab:human_corr}
  \begin{tabular*}{\columnwidth}{@{}l@{\extracolsep{\fill}}lcccc@{}}
    \toprule
    \textbf{Dimension} & \textbf{Sub-metric} & \textbf{SRCC} & \textbf{KRCC} & \textbf{PLCC} & \textbf{Model} \\
    \midrule
    \multirow{2}{*}{\makecell[l]{Visual\\Quality}} & AS             & 46.0 & 32.9 & 49.2 & 50.8 \\
                             & PS             & 39.7 & 28.5 & 46.4 & 30.5 \\
    \midrule
    \multirow{2}{*}{\makecell[l]{Dynamic\\Quality}} & DD$^\dagger$   & 60.7 & 44.0 & 60.0 & 69.4 \\
                             & PP             & 30.5 & 20.8 & 28.1 & 66.2 \\
    \midrule
    \multirow{4}{*}{\makecell[l]{Semantic\\Alignment}} & VQA            & 64.6 & 47.7 & 66.5 & 85.5 \\
                             & TC             & 89.6 & 79.6 & 92.0 & 95.3 \\
                             & CTA            & 81.7 & 67.2 & 78.5 & 97.7 \\
                             & CC             & 66.1 & 48.7 & 68.7 & 87.3 \\
    \midrule
    \multirow{3}{*}{\makecell[l]{Cross-Shot\\Consistency}} & CSEC           & 46.3 & 34.2 & 60.2 & 61.0 \\
                             & CSBC           & 39.4 & 28.9 & 46.4 & 84.7 \\
                             & CSFC           & 63.2 & 47.8 & 48.4 & 68.1 \\
 
    \bottomrule
  \end{tabular*}
\end{table}

\paragraph{Agreement.}
Table~\ref{tab:human_corr} reports the agreement of every sub-metric with the mean human rating. The structural metrics (TC, CTA) agree best, followed by Camera Control, VQA-Score, Face Consistency, and cross-validated DD. The appearance-oriented items (AS, PS, CSEC, CSBC) reach only moderate agreement: aesthetic and appearance judgments vary noticeably from person to person, which is also reflected in their lower inter-annotator ICC. For Visual and Dynamic Quality, we compared several off-the-shelf scorers and keep the best-agreeing ones (Appendix~\ref{app:candidates}). Physical Plausibility is the weakest at the video level but still separates systems well (Appendix~\ref{app:candidates}). At the system level, which is what a leaderboard reports, agreement is higher for most sub-metrics. Finally, the system ranking by the Final Score correlates with the ranking fitted to the pairwise overall preferences with SRCC \numFinalBT{} (Appendix~\ref{app:human_correlation}), showing that \name{} ranks systems consistently with human judgment.

\paragraph{Judge robustness.}
Four sub-metrics (PP, VQA, CSEC, CSBC) are scored by Qwen3.8, which raises the question of whether a different VLM judge would lead to different conclusions. To test this, we re-score these sub-metrics on all videos with Doubao-Seed-2.1-lite, a VLM from a different model family, keeping prompts and inputs fixed. The system rankings by the Final Score under the two judges correlate with SRCC \numJudgeFinal{}, and the system rankings by the four sub-metrics with SRCC \numJudgeModLo{}--\numJudgeModHi{} (Appendix~\ref{app:judge_robustness}). The choice of VLM judge therefore has little effect on the system ranking, although single-video scores of the weaker sub-metrics, Physical Plausibility in particular, depend more on the judge.

\section{Conclusion}
\label{sec:conclusion}

We presented \name{}, a benchmark for controllable text-driven multi-shot video generation whose Scene-Entity-Shot scripts specify the timing, camera state, content, and transition of every shot, evaluated by 11 sub-metrics validated against human ratings. Across 14 systems, visual quality does little to separate the systems, whereas many of them cut mainly on scene changes, render gradual transitions as hard cuts, fall back to common camera states, and lose structure on longer scripts. The resulting system ranking agrees with human judgment (system-level SRCC \numFinalBT{}) and changes little under a different VLM judge, so \name{} offers a reliable testbed for tracking whether future systems close these gaps. Future models will need explicit shot-level conditioning and training data with within-scene cuts, labeled transitions, and long-tail camera states, and multi-shot benchmarks should report structural validity separately from quality.

{
  \small
  \bibliographystyle{ieeenat_fullname}
  \bibliography{reference}
}

\appendix
\renewcommand{\topfraction}{0.9}\renewcommand{\dbltopfraction}{0.9}
\renewcommand{\textfraction}{0.07}
\renewcommand{\floatpagefraction}{0.8}\renewcommand{\dblfloatpagefraction}{0.8}
\par\ifdim\dimexpr\pagegoal-\pagetotal\relax<8\baselineskip\newpage\else\vspace{1.5em}\fi
{\centering\Large\textbf{Appendix}\par}
\vspace{0.5em}
\noindent The benchmark scripts (300 test cases), prompts, evaluation toolkit, reproduction guidelines, and example videos generated by representative models are available on the project page, \projecturl{}.

\section{Limitations and Future Directions}
\label{app:limitations}

\name{} evaluates text-driven multi-shot generation on the visual stream. Some recent systems also take reference images of characters or scenes as input, or generate audio together with the video~\citep{wei2026msavbench}, and these settings are beyond the scope of this work. Since every sub-metric is computed per shot from the structured script, metrics for reference conditioning or audio-visual synchronization can later be added as new sub-metrics without changing the rest of the pipeline.

Several semantic sub-metrics rely on a VLM judge. If the judge were a closed-source API model, it could be updated or retired at any time, and earlier scores would no longer be reproducible. We therefore use Qwen3.8-27B, whose weights are public. Although our runs call it through a hosted API, the same checkpoint can be deployed locally, and its scores are read as token-probability expectations rather than sampled answers (Appendix~\ref{app:vlm_judge}). Replacing it with a judge from a different model family changes the system ranking only slightly (Section~\ref{sec:human} and Appendix~\ref{app:judge_robustness}), and stronger open-weight judges can take its place as they become available.

Finally, commercial systems are accessed through their public APIs and may be updated after our evaluation, so the reported scores reflect the versions available at that time. We release the scripts and the evaluation pipeline on the project page, which allows new versions to be evaluated under the same protocol.

\section{Dataset Details}
\label{app:dataset_details}

\begin{figure*}[!t]
  \centering
  \begin{subfigure}[b]{0.355\textwidth}
    \centering
    \includegraphics[width=\linewidth]{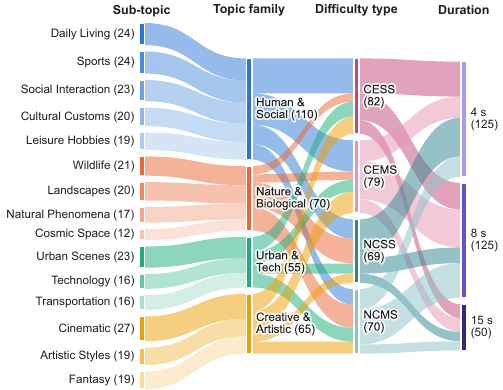}
    \caption{Script flow (script counts).}
    \label{fig:app_sankey}
  \end{subfigure}\hfill
  \begin{subfigure}[b]{0.625\textwidth}
    \centering
    \includegraphics[width=\linewidth]{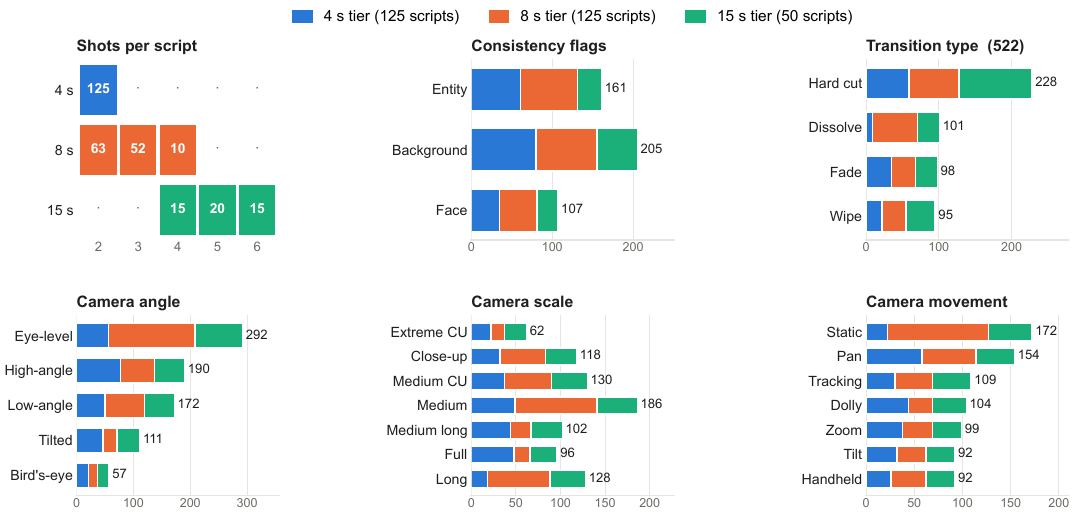}
    \caption{Shot-level statistics, stacked by duration tier.}
    \label{fig:app_shot_stats}
  \end{subfigure}
  \caption{Distribution of the 300 scripts.}
  \label{fig:dataset_distribution}
\end{figure*}

\paragraph{Headline statistics.}
\name{} contains 300 scripts, 822 shots, and 522 transitions. The 4\,s tier contains 125 scripts, 250 shots, and 125 transitions; the 8\,s tier contains 125 scripts, 322 shots, and 197 transitions (63, 52, and 10 scripts with two, three, and four shots); the 15\,s extension contains 50 scripts, 250 shots, and 200 transitions (15, 20, and 15 scripts with four, five, and six shots). The four difficulty types are distributed as CrossEntitySingleScene (CESS) 82, CrossEntityMultiScene (CEMS) 79, NocrossEntityMultiScene (NCMS) 70, and NocrossEntitySingleScene (NCSS) 69. The scripts register 585 entities, 300 of them human, and contain 6{,}127 VQA questions.

\paragraph{Script schema.}
Each script is a JSON record with the fields in Table~\ref{tab:schema}. The text prompt given to every system is rendered from the global context, the entity registry, and the shot sequence, so it contains every requested subject, scene, camera state, timestamp, and transition.

\begin{table}[!htbp]
  \centering
  \caption{Fields of a \name{} script.}
  \label{tab:schema}
  \footnotesize
  \setlength{\tabcolsep}{3pt}
  \begin{tabular}{@{}>{\raggedright\arraybackslash}p{2.7cm}>{\raggedright\arraybackslash}p{\dimexpr\linewidth-2.7cm-2\tabcolsep\relax}@{}}
    \toprule
    \textbf{Field} & \textbf{Content} \\
    \midrule
    \texttt{type}, \texttt{topic} & difficulty type (cross-entity or not, single or multiple scenes) and topic \\
    \texttt{global\_context} & setting, lighting, and atmosphere, with \texttt{[SceneK]} tags for each scene \\
    \texttt{entities\_info} & registered entities: \texttt{id}, appearance \texttt{caption}, \texttt{is\_human} \\
    \texttt{shot\_sequence} & per shot: start and end time, camera angle, scale, and movement, description referring to \texttt{[entityK]} and \texttt{[SceneK]}, and the transition to the next shot (type, duration, description) \\
    \texttt{VQA\_questions} & yes/no questions with the \texttt{shot\_ids} where the answer should be visible \\
    \texttt{eval\_metrics} & which of entity, background, and face consistency apply \\
    \texttt{background\_group} & groups of shots that share a scene \\
    \texttt{duration\_s} & duration tier (4, 8, or 15\,s) \\
    \texttt{t2v\_prompt} & the text prompt rendered from the fields above \\
    \bottomrule
  \end{tabular}
\end{table}

\paragraph{Distribution design.}
Figure~\ref{fig:app_sankey} traces every script from its sub-topic and topic family to its difficulty type and duration tier; numbers are script counts. The dataset is intentionally not uniform over topics, difficulty types, camera states, and transitions. Some topics have a naturally narrow visual vocabulary: Cosmic Space prompts, for example, easily collapse into repeated rockets, planets, astronauts, and star fields if oversampled. In contrast, Daily Living can support many concrete scenarios. We therefore balance coverage against diversity within each topic rather than enforcing equal counts; sub-topics range from 27 scripts (Cinematic) to 12 (Cosmic Space).

\paragraph{Transition and camera distributions.}
Figure~\ref{fig:app_shot_stats} gives the shot-level statistics: shots per script, the number of scripts on which each cross-shot consistency metric is evaluated, and the requested transition types and camera states, with bars stacked by duration tier and the total at the bar end. Transitions comprise Hard Cut 228, Dissolve 101, Fade 98, and Wipe 95. Camera angles are Eye-level 292, High-angle 190, Low-angle 172, Tilted 111, and Bird's-eye-view 57. Camera scales are Medium 186, Medium-close-up 130, Long 128, Close-up 118, Medium-long 102, Full 96, and Extreme-close-up 62. Camera movements are Static 172, Pan 154, Tracking 109, Dolly 104, Zoom 99, Tilt 92, and Handheld 92.

\paragraph{Cross-shot metric coverage.}
Each script is labeled with the cross-shot consistency metrics that apply to it: Entity Consistency applies to 161 scripts, Background Consistency to 205, and Face Consistency to 107. These labels were generated during construction and manually reviewed.

\section{Evaluation Metric Details}
\label{app:metric_details}

This section specifies how each sub-metric is computed in the released evaluation code. Appendix~\ref{app:transition_validation} describes the shot segmentation shared by all shot-level metrics, and the remaining subsections give, for each sub-metric, its frame sampling, how shot scores are combined into a video score, and the scoring rule. All sub-metrics are mapped to $[0,1]$ before aggregation (Appendix~\ref{app:aggregation}). The VLM prompt templates are released verbatim with the evaluation code.

\begin{table}[t]
  \centering
  \caption{Overview of the \name{} sub-metrics and their underlying models or tools. All VLM-based evaluations use Qwen3.8-27B. Implementation details follow in this section.}
  \label{tab:metric_overview}
  \scriptsize
  \setlength{\tabcolsep}{2pt}
  \renewcommand{\arraystretch}{1.1}
  \begin{tabular}{@{}>{\raggedright\arraybackslash}m{0.155\columnwidth}>{\raggedright\arraybackslash}m{0.40\columnwidth}>{\raggedright\arraybackslash}m{0.405\columnwidth}@{}}
    \toprule
    \textbf{Dimension} & \textbf{Sub-metric} & \textbf{Underlying Model / Tool} \\
    \midrule
    \multirow{2}{=}{Visual Quality}
      & Aesthetic Score (AS)            & Q-Align~\citep{wu2023qalign} \\
      & Perceptual Score (PS)           & DeQA-Score~\citep{deqa_score} \\
    \midrule
    \multirow{2}{=}{Dynamic Quality}
      & Dynamic Degree (DD)             & DEVIL~\citep{liao2024devil} \\
      & Physical Plausibility (PP)      & Qwen3.8~\citep{qwen38} \\
    \midrule
    \multirow{4}{=}{Semantic Alignment}
      & VQA-Score (VQA)                 & Qwen3.8 yes/no question answering \\
      & Transition Control (TC)         & OmniShotCut~\citep{wang2026omnishotcut} + TransVLM~\citep{chen2026transvlm} + Qwen3.8 \\
      & Cut Timing Accuracy (CTA)       & Exp-penalized shot timing error \\
      & Camera Control (CC)             & FilmOps~\citep{wang2026filmbench} cinematography operators \\
    \midrule
    \multirow{3}{=}{Cross-Shot Consistency}
      & Entity Consistency (CSEC)       & Qwen3.8 \\
      & Background Consistency (CSBC)   & Qwen3.8 \\
      & Face Consistency (CSFC)         & InsightFace~\citep{deng2019arcface} + Qwen3.8 identity selection \\
    \bottomrule
  \end{tabular}
\end{table}

\subsection{Shot Segmentation and Transition Recognition}
\label{app:transition_validation}

Transition detection underlies most of the evaluation. The detected events are the direct input of TC and CTA, and the shot intervals they induce are used by every shot-level metric, so a missed or mislocalized transition propagates to the sub-metrics of all four dimensions. Detectors used in prior work, such as PySceneDetect~\citep{castellano_pyscenedetect} and TransNet~V2~\citep{soucek2024transnet}, are designed mainly for abrupt cuts and in our tests often miss gradual transitions or localize them poorly, whereas 294 of the 522 transitions requested by \name{} are dissolves, fades, or wipes. We therefore combine two recent detectors: OmniShotCut~\citep{wang2026omnishotcut} for hard cuts and TransVLM~\citep{chen2026transvlm} for gradual transitions, with Qwen3.8 as an auxiliary judge for cases the detectors leave open.

\paragraph{Pipeline.}
Hard cuts are taken from OmniShotCut as time points, and gradual-transition intervals from TransVLM, discarding an interval that merely covers a detected hard cut. An interval inherits the type of the OmniShotCut dissolve, fade, or wipe segment it overlaps most; otherwise Qwen3.8 classifies it from the full video with the interval location given in the prompt, and the interval is dropped if Qwen3.8 types it as a hard cut or answers \emph{none} in a second query that also offers \emph{other} and \emph{none}. These rules were fixed after a manual review of detections on a separate set of 151 videos. For frame-based metrics (AS, PS, DD, CC), two frames on each side of a hard cut and all gradual-transition frames are excluded from the shots; for VQA slots and face sampling, shots are shrunk by 0.1\,s at each boundary and shots shorter than 0.3\,s are discarded. All systems, including VGoT, are segmented by the same pipeline.

\paragraph{Diagnostic set.}
To choose the pipeline, we manually annotated a diagnostic set of 150 videos generated by 11 systems from 4\,s and 8\,s scripts, containing 135 transition events (59 hard cuts, 26 dissolves, 25 fades, and 25 wipes). Predictions are matched one-to-one within each video: a hard cut must be predicted within $\pm$0.2\,s, and a gradual transition as an interval with temporal IoU $\geq$0.5; typed F1 additionally requires the correct type.

\begin{table}[!htbp]
  \centering
  \caption{Transition detection on the diagnostic set (150 videos, 135 events). Hard and Gradual are localization F1 restricted to hard cuts and to gradual transitions.}
  \label{tab:transition_validation}
  \footnotesize
  \setlength{\tabcolsep}{1.5pt}
  \begin{tabular*}{\linewidth}{@{\extracolsep{\fill}}lcccc@{}}
    \toprule
    & \multicolumn{4}{c}{\textbf{F1}} \\
    \cmidrule(l){2-5}
    \textbf{Pipeline} & \textbf{Loc.} & \textbf{Typed} & \textbf{Hard} & \textbf{Grad.} \\
    \midrule
    PySceneDetect only (all as hard cuts) & 0.475 & 0.475 & 0.700 & 0.000 \\
    TransNet~V2 only (intervals untyped) & 0.645 & 0.473 & 0.866 & 0.378 \\
    PySceneDetect + VLM refining/typing & 0.794 & 0.719 & 0.920 & 0.701 \\
    OmniShotCut only (native labels) & 0.806 & 0.761 & 0.899 & 0.719 \\
    TransVLM only + Qwen3.8 typing & 0.861 & 0.810 & 0.883 & 0.844 \\
    OmniShotCut + TransVLM, Qwen3.8 typing & 0.889 & 0.824 & 0.936 & 0.852 \\
    OmniShotCut + TransVLM, fused typing & 0.889 & 0.853 & 0.936 & 0.852 \\
    \textbf{Ours} (fused typing + removal) & \textbf{0.900} & \textbf{0.885} & \textbf{0.951} & \textbf{0.857} \\
    \bottomrule
  \end{tabular*}
\end{table}

As Table~\ref{tab:transition_validation} shows, the conventional detectors find most hard cuts but few gradual transitions: PySceneDetect reports every event as a hard cut and reaches a Gradual F1 of 0, and TransNet~V2 (official weights, default threshold 0.5) reaches 0.378 even when its many-hot head is used to form intervals. OmniShotCut alone is accurate on hard cuts but reaches only 0.719 on gradual transitions; adding TransVLM raises the Gradual F1 to 0.852 and the localization F1 to 0.889, and fused typing with the removal rules raises the typed F1 from 0.824 to 0.885. The final pipeline types 58/59 hard cuts, 22/26 dissolves, 21/25 fades, and 18/25 wipes correctly, so wipes remain the hardest class. Preferring OmniShotCut's native label was decided after inspecting this set; without it, the typed F1 is 0.824. The remaining errors have little effect on the scores: TC computed from the detected events equals TC from the annotated events on 143 of the 150 videos (95.3\%), with SRCC 0.901 at the video level and 0.927 at the model level, and the corresponding SRCC for CTA are 0.874 and 0.936.

\paragraph{Held-out check.}
After freezing the pipeline, we checked it on 84 further videos covering what the diagnostic set leaves out: six 15\,s videos from each of the nine systems evaluated on the extension, and five 4\,s and five 8\,s videos from each of the three systems absent from the diagnostic set (Kling~3.0, Seedance~2.0, and MiniMax~H3). An annotator corrected the pipeline's events frame by frame with the system name hidden, which may favor the pipeline. On the resulting 218 events, the pipeline reaches a localization F1 of 0.939 and a typed F1 of 0.930 (0.931 on the 15\,s videos and 0.925 on the new systems), and TC from the detected events matches TC from the corrected events on 89.3\% of the videos.

The 144 clips of the scene-change intervention (Appendix~\ref{app:c1_intervention}) were labeled blind to the system and to any detection, with the number of shots in the first 4\,s and the transition type. On them, the pipeline finds the labeled shot count in 132 clips (91.7\%; Cohen's $\kappa=0.847$ on whether a boundary exists) and assigns the correct family (hard cut or gradual) to 57 of the 59 matched boundaries. Recomputing the intervention from the detected events gives the same conclusion: boundaries rise from 21 to 47 of the 72 clips per version, increasing in 17 scripts and decreasing in one ($p=8\times10^{-5}$).

\subsection{Visual Quality}
\label{app:visual_details}

\paragraph{Aesthetic Score.}
Q-Align~\citep{wu2023qalign} is queried with its official video-quality prompt on the frames of each clean shot sampled at 4\,fps, treated as one clip. As in the official scorer, the score is read from the probabilities of the five level words at the answer position:
\begin{equation}
\mathrm{AS}_{\mathrm{shot}}=\sum_{\ell} p_\ell\, v_\ell,
\end{equation}
where $p_\ell$ is the softmax over the five level tokens and $(v_{\text{excellent}},v_{\text{good}},v_{\text{fair}},v_{\text{poor}},v_{\text{bad}})=(1,0.75,0.5,0.25,0)$. Shot scores are averaged with weights proportional to shot duration, so that a short shot does not count as much as a long one.

\paragraph{Perceptual Score.}
DeQA-Score~\citep{deqa_score} rates each frame sampled at 8\,fps within a clean shot. Frame scores are averaged into a shot score, shot scores are averaged with duration weights, and the result is divided by 5 and clipped to $[0,1]$.

\subsection{Dynamic Quality}
\label{app:dynamic_details}

\paragraph{Dynamic Degree.}
Within each clean shot of at least 0.4\,s, we compute the seven DEVIL~\citep{liao2024devil} dynamics components at 8\,fps (optical-flow magnitude, SSIM change, perceptual-hash change, segment-level change of DINO and ViCLIP features, DINO information change, and temporal entropy), so that a cut is not counted as motion. Each component is averaged over the shots of a video and standardized over all 3{,}950 benchmark videos. The official DEVIL fusion was trained on single-shot clips and agrees only moderately with our annotators (Appendix~\ref{app:candidates}), so we map the standardized components $\mathbf{z}$ to the 1--5 human motion scale with a ridge regression ($\lambda=1$, fixed in advance), $\mathrm{DD}=b+\mathbf{w}^{\top}\mathbf{z}$, fit on all 218 human-rated videos. To avoid overstating agreement, the SRCC reported for DD is cross-validated (5 folds grouped by script, averaged over 100 random partitions): \numDDcvPrecise{} at the video level and \numDDcvModelPrecise{} at the model level, insensitive to $\lambda\in[0.01,10]$.

\paragraph{Physical Plausibility.}
PP is judged once per video rather than per shot: Qwen3.8 watches the whole video, sampled at 10\,fps, and goes through a failure checklist shot by shot: (A) body deformation, (B) object continuity, (C) motion quality, and (D) common sense and gravity. It lists the visible failures under each item and then gives a 1--5 rating whose anchors follow the human rubric. A free-form rating saturates quickly, because most generated videos look plausible at first glance; the checklist therefore names the failures annotators actually penalize, including secondary elements such as cloth, water, or smoke that move on their own and hands or tools that never make contact, and reserves the top rating for videos in which every item is clean. Real cuts and transitions, prompt adherence, stylization, and image-quality defects are excluded from the judgment, and a subject that should move but stays frozen counts as a motion failure. The script is not shown, so the rating cannot be influenced by what the video was supposed to contain.

\subsection{Semantic Alignment}
\label{app:semantic_metrics}

\paragraph{VQA-Score.}
Each question lists the shots in which its answer should be visible. Each requested shot is mapped to the detected shot with the same index, which defines a time window (slot). Every question is asked once for each of its slots and once for the whole video. All (question, slot) pairs of a video are answered in a single request on the whole video at 4\,fps: each pair is one numbered line, and a slot is written into the line as ``Judge only the time window $a$\,s to $b$\,s of the video.'' Asking on the whole video, rather than on cut clips, keeps the model's view of the video identical across questions. Let $p_1,\dots,p_k$ be the yes-probabilities on the $k$ slots of a question and $p_G$ the yes-probability on the whole video. A requested shot without a detected counterpart contributes $p=0$; slots shorter than 1\,s are dropped. With $n=k+1$,
\begin{equation}
s_q=\frac{1}{n}\Big(\sum_{i=1}^{k}p_i+\max\big(p_G,\max_i p_i\big)\Big).
\end{equation}
When probabilities are 0 or 1, this reduces to full credit when the answer is yes in every requested shot and $1/n$ when it is yes only on the whole video, so an event that occurs in the wrong shot earns little credit. Questions without shot IDs use $p_G$. The VQA-Score averages $s_q$ over the questions of a video.

The index-based mapping matters only when the detected shot count differs from the requested one. As a check, we instead map requested shots to detected shots by the monotone one-to-one alignment that maximizes the summed temporal IoU (described under Camera Control below); a requested shot left without a counterpart counts as out of range. This changes the slots of 642 videos, of which 636 are re-queried with the same prompt and settings, and the remaining six reuse the original answers. The base-set VQA-Score of every system changes by at most 0.9 points, the ranking of the 14 systems is unchanged, and the video-level SRCC with human ratings is 0.640 vs.\ 0.646 for the index-based mapping.

\paragraph{Transition Control.}
Let $\mathcal{T}=(t_1,\dots,t_n)$ be the requested transitions, ordered in time (a hard cut is a point at the end of its shot; a gradual transition spans its requested duration), and $\hat{\mathcal{T}}=(\hat{t}_1,\dots,\hat{t}_m)$ the detected events. We select $\min(n,m)$ pairs $(i_k,j_k)$ with $i_1<\dots<i_K$ and $j_1<\dots<j_K$ that minimize $\sum_k |c(t_{i_k})-c(\hat{t}_{j_k})|$, where $c(\cdot)$ is the event center, by dynamic programming. Then
\begin{equation}
\begin{gathered}
\mathrm{TC\text{-}V}=\frac{K}{n},\qquad \mathrm{TC}=\mathrm{TC\text{-}V}\cdot\mathrm{TC\text{-}A},\\
\mathrm{TC\text{-}A}=\frac{1}{K}\sum_{k}\mathbb{I}\big[\mathrm{type}(t_{i_k})=\mathrm{type}(\hat{t}_{j_k})\big].
\end{gathered}
\end{equation}
TC-V is the share of requested transitions that are rendered at all, and TC-A the share of rendered ones that have the requested type, so a missing transition lowers only TC-V and a wrongly typed one only TC-A. TC-A is undefined when $K=0$ and is averaged over videos with $K\ge1$. The SA score uses TC, the share of requested transitions rendered with the requested type.
TC and CTA use only the detected events and shot intervals, without further frame sampling. Extra detected events are ignored and unmatched requests count as failures. The order constraint prevents a spurious early cut from shifting all later matches; no distance cap is used, because timing errors are already penalized by CTA. The other shot-level metrics (CTA, VQA, and CC) instead pair the $i$-th requested shot with the $i$-th detected shot, so that an extra or missing shot is penalized rather than skipped; the two matchings coincide whenever the numbers of requested and detected events are equal. Scripts without requested transitions have no TC.

\paragraph{Cut Timing Accuracy.}
CTA compares expected and detected shot intervals in seconds. Similar metrics in HoloCine~\citep{meng2025holocine} and ShotAdapter~\citep{kara2025shotadapter} count deviations in frames, which penalizes systems differently when they render at different frame rates; CTA is independent of the frame rate and also covers the durations of gradual transitions. Let $m_s=\min(N_s,\hat{N}_s)$ be the number of shots matched in order, with expected boundaries $(s_i,e_i)$ and detected boundaries $(\hat{s}_i,\hat{e}_i)$. Extra detected shot durations are denoted $\hat{d}^{+}_j$ and missing expected shot durations $d^{-}_j$. The shot-timing error is
\begin{equation}
\begin{split}
E_{\mathrm{shot}} ={}& \sum_{i=1}^{m_s}\left(|\hat{s}_i-s_i|+|\hat{e}_i-e_i|\right)\\
&+\sum_j (j+1)\hat{d}^{+}_j+\sum_j (j+1)d^{-}_j .
\end{split}
\end{equation}
The same increasing penalty is applied to transition durations. For matched transitions with durations $\delta_i$ and $\hat{\delta}_i$, extra detected transition durations $\hat{u}^{+}_j$, and missing expected transition durations $u^{-}_j$,
\begin{equation}
E_{\mathrm{trans}} =
\sum_{i=1}^{m_t}|\hat{\delta}_i-\delta_i|
+\sum_j (j+1)\hat{u}^{+}_j
+\sum_j (j+1)u^{-}_j .
\end{equation}
The score is normalized by the video duration $T$:
\begin{equation}
\mathrm{CTA}=\exp\!\left(-\frac{E_{\mathrm{shot}}+E_{\mathrm{trans}}}{T}\right).
\end{equation}
Matched boundaries are penalized linearly, while each further extra or missing segment receives an increasing multiplier ($2,3,\ldots$): one extra or missing shot can still read as a local timing error, but several of them change the structure of the video. CTA is computed for every system, including those that receive per-shot lengths as input (HoloCine, MSM, and VGoT): the benchmark scores the delivered video against the script, whatever pipeline produced it. Receiving the shot lengths does not guarantee a high score. VGoT renders each shot separately and concatenates the clips, but its shot generator supports only a discrete set of clip lengths: the requested shot lengths are allocated proportionally within the total frame budget and mapped to supported lengths, which leaves errors at the shot boundaries. Its CTA is therefore lower than that of several systems that generate the whole video at once (Table~\ref{tab:main_results}).

\paragraph{Camera Control.}
FilmOps~\citep{wang2026filmbench} predicts shot scale and camera angle on frames sampled at 2\,fps within each clean shot and camera movement from the whole shot clip. The $i$-th requested shot is compared with the $i$-th detected shot. FilmOps labels are mapped to the \name{} vocabulary with partial credit for neighboring classes, fixed before any comparison with human ratings: a one-step scale difference earns 0.5; an eye-level prediction earns 0.5 for a requested high or low angle; a zoom earns 0.5 for a requested dolly and a push-in or pull-out earns 0.5 for a requested zoom; a tilted angle requires a Dutch-angle prediction. Let $g_a(\cdot,\cdot)\in\{0,0.25,0.5,0.75,1\}$ be this similarity for aspect $a\in\{\text{scale},\text{angle},\text{movement}\}$, $c_{i,a}$ the requested value for shot $i$, $\mathcal{K}_i$ the frames of detected shot $i$ with a known FilmOps label, and $N$ the number of requested shots. Scale and angle ($a\in\{\text{scale},\text{angle}\}$) are averaged over the frames of each shot, and movement ($m$) is scored once per shot:
\begin{equation}
\begin{gathered}
\mathrm{CC}_a=\frac{1}{N}\sum_{i=1}^{N}\frac{1}{|\mathcal{K}_i|}\sum_{f\in\mathcal{K}_i} g_a(\hat{c}_{f,a},c_{i,a}),\\
\mathrm{CC}_{m}=\frac{1}{N}\sum_{i=1}^{N} g_{m}(\hat{c}_{i,m},c_{i,m}),
\end{gathered}
\end{equation}
\begin{equation}
\mathrm{CC}=\tfrac{1}{3}\big(\mathrm{CC}_{\text{scale}}+\mathrm{CC}_{\text{angle}}+\mathrm{CC}_{m}\big),
\end{equation}
where a requested shot without a detected counterpart, or without any known label, contributes 0. FilmOps may return several angle or movement labels for one input; the best-matching label is used. CC therefore reflects both shot structure and camera execution. To separate the two, we also report CC-V, the share of the $N$ requested shots that have a detected counterpart with a valid FilmOps output.

The order-based correspondence matters only when the detected shot count differs from the requested one, which holds for 33\% of the base-set videos. Aligning planned and detected shots instead by a monotone one-to-one assignment that maximizes the summed temporal IoU changes the system-level CC by at most 2.6 points (SRCC 0.99 and Kendall's $\tau$ 0.96 with the order-based ranking over the 14 systems) and leaves the agreement with human ratings unchanged (video-level SRCC 0.666 vs.\ 0.661 on the 218 rated videos). Assigning each planned shot the detected shot with the largest overlap, without the one-to-one constraint, lets a single long shot stand in for several requested shots: it raises the CC of systems that rarely cut by up to 17.0 points and agrees less with the annotators (SRCC 0.562).

\subsection{Cross-Shot Consistency}
\label{app:csc_details}

\paragraph{Applicability.}
Each consistency metric is computed only on scripts labeled for it (Appendix~\ref{app:dataset_details}). Scripts with no enabled relation receive no CSC score; at the system level, CSC averages the three consistency sub-metrics over the scripts that enable each of them (Appendix~\ref{app:aggregation}).

\paragraph{Entity and background consistency.}
Qwen3.8 receives the whole video at 10\,fps and the script, which serves only to identify the entities and scenes meant to recur. For CSEC, the prompt lists the registered entities with their captions; the judge matches each entity across the shots where it appears, lists its visible cross-shot changes, and rates consistency of shape, body proportions, colors, materials, clothing, patterns, and key parts. Differences explained by viewpoint, pose, occlusion, lighting, or a state change requested by the script are not penalized, and face identity is left to CSFC. For CSBC, the prompt states the scene assigned to each shot; the judge compares only shots intended to share an environment and rates the stability of scene identity, fixed structures, spatial layout, materials and textures, and landmarks. Shots of different intended scenes are not compared, and viewpoint, shot-scale, and lighting differences, moving objects, and foreground subjects are not penalized by themselves. An applicable video is valid for CSEC and CSBC (CSEC-V, CSBC-V) when at least two shots are detected; an invalid video counts as 0 (Appendix~\ref{app:aggregation}), since a single-shot video would otherwise be rated as highly consistent. Among valid videos there is no separate check that the relation appears: if a recurring entity or scene is missing, the rating drops accordingly.

\paragraph{Face consistency.}
\label{app:csfc}
CSFC is computed on scripts labeled for face consistency that register at least one human entity. InsightFace (\texttt{buffalo\_l})~\citep{deng2019arcface} detects all faces on full-resolution frames sampled at 4\,fps inside each shot slot. For each human entity, up to six frames with faces per shot are sent to Qwen3.8 with all detected faces drawn as numbered boxes, the entity caption, and the captions of the other people in the script; the judge returns the box number of the target or 0 if the target is absent (Set-of-Mark prompting, greedy decoding). The identity features of the selected faces are grouped by shot and compared by cosine similarity. Raw similarities between ArcFace features are low even for the same person: videos that annotators rate as fully consistent (mean rating of at least 4.5) obtain only 51.5 on average from raw similarities, far below the range of the other sub-metrics. We therefore map each similarity with the monotonic function
\begin{equation}
f(x)=1-(1-x)^{\beta}, \qquad \beta=\ln(0.4)/\ln(0.7)\approx2.569,
\end{equation}
which keeps $f(0)=0$ and $f(1)=1$ and is anchored at a commonly used verification threshold, mapping 0.3 to 0.6 and 0.6 to 0.9. Because $f$ is monotonic, it leaves the agreement with human ratings essentially unchanged (SRCC 0.632, against 0.633 without calibration) and only brings CSFC onto a scale comparable with the other sub-metrics.

The score is not the average of all cross-shot feature pairs. The same character can appear with different face sizes, head poses, crop qualities, and occlusions across shots, and averaging all pairs would be dominated by low similarities between incompatible poses even when both shots show the same identity. We therefore use nearest-neighbor matching: each selected face is compared with the most similar face of the other shot. Let $\mathcal{F}_{e,a}$ be the L2-normalized features of the selected faces of entity $e$ in shot $a$. For two shots $a$ and $b$ that both contain selected faces, the directional score is
\begin{equation}
r_{a\rightarrow b}=
\frac{1}{|\mathcal{F}_{e,a}|}
\sum_{z\in\mathcal{F}_{e,a}}
f\!\left(\max_{z'\in\mathcal{F}_{e,b}}\max(0,z^{\top}z')\right),
\end{equation}
and the shot-pair score is $r_{a,b}=\tfrac12(r_{a\rightarrow b}+r_{b\rightarrow a})$. For an entity found in the set of shots $\mathcal{S}_e$,
\begin{equation}
\mathrm{CSFC}_e=
\frac{2}{|\mathcal{S}_e|(|\mathcal{S}_e|-1)}
\sum_{\substack{a,b\in\mathcal{S}_e\\a<b}} r_{a,b}.
\end{equation}
The video-level CSFC averages $\mathrm{CSFC}_e$ over human entities found in at least two shots. If no entity is found in two shots, the video receives 0: in the human set, applicable videos whose person could not be re-identified across shots were consistently rated low.

\subsection{VLM Judge Settings}
\label{app:vlm_judge}
All VLM judgments (transition typing, PP, VQA, CSEC, CSBC, and face selection) use the 27B checkpoint of Qwen3.8~\citep{qwen38}, accessed through the Alibaba Cloud Model Studio API (\texttt{qwen3.8-27b}) on full-resolution videos. Scoring requests disable thinking and use the model card's recommended non-thinking sampling (temperature 0.7, top-$p$ 0.8, top-$k$ 20, presence penalty 1.5) with a fixed seed (2027) and top-5 log-probabilities. Transition typing enables thinking with low reasoning effort and uses temperature 1.0, top-$p$ 0.95, and top-$k$ 20 with the same seed, and face selection uses greedy decoding with a 16-token answer limit.

Sampling affects only the free-text evidence that precedes the answer. Rubric scores (PP, CSEC, CSBC) are not read from the sampled digit: we take the probability-weighted mean over the tokens 1--5 at the position of the score,
\begin{equation}
\hat{s}=\frac{\sum_{d=1}^{5} d\,P(d)}{\sum_{d=1}^{5}P(d)},
\end{equation}
and VQA uses $P(\text{yes})/(P(\text{yes})+P(\text{no}))$ at each answer token. Both are more stable across repeated requests than sampled answers and resolve ties among videos that would receive the same integer score.

\subsection{Hierarchical Aggregation}
\label{app:aggregation}

Every sub-metric is mapped to $[0,1]$: 1--5 scores (PP, CSEC, CSBC, and DD) via $(s-1)/4$ with clipping, DeQA scores by dividing by 5, and all other sub-metrics are already in $[0,1]$. Scores are aggregated per system and split in three levels. First, each sub-metric is averaged over the videos to which it applies; for the validity-gated consistency sub-metrics this equals the validity times the score, \eg, $\mathrm{CSEC}=\mathrm{CSEC\text{-}V}\cdot\mathrm{CSEC\text{-}S}$, since an invalid video counts as 0. For Transition Control, whose validity is a per-video ratio, the system-level score is the product of the two system-level averages, $\mathrm{TC}=\mathrm{TC\text{-}V}\cdot\mathrm{TC\text{-}A}$, which differs from the mean of the per-video TC by at most 0.7 points on the base set. Second, within a dimension, these sub-metric scores are averaged with equal weights, skipping any that does not apply:
\begin{equation}
D_k=\frac{\sum_{m\in\mathcal{M}_k} s_{k,m}\,\mathbb{I}[s_{k,m}\neq\varnothing]}
{\sum_{m\in\mathcal{M}_k} \mathbb{I}[s_{k,m}\neq\varnothing]},
\end{equation}
where $\varnothing$ marks an inapplicable sub-metric. Visual Quality averages AS and PS, Dynamic Quality averages DD and PP, Semantic Alignment averages VQA, TC, CTA, and CC, and Cross-Shot Consistency averages CSEC, CSBC, and CSFC.

Third, the Final Score is the equal-weight mean of the four dimension scores,
\begin{equation}
\mathrm{FinalScore}=\frac{\sum_{k} D_k\,\mathbb{I}[D_k\neq\varnothing]}{\sum_{k} \mathbb{I}[D_k\neq\varnothing]},
\label{eq:final_score}
\end{equation}
with $k\in\{\mathrm{VQ},\mathrm{DQ},\mathrm{SA},\mathrm{CSC}\}$, so that every sub-metric carries the same weight within its dimension and every dimension contributes exactly $1/4$, regardless of how many scripts enable each consistency relation, and the Final Score in the result tables is the mean of the four dimension columns. For video-level analyses (bootstrap resampling is done on scripts and re-aggregated as above; the pairwise check below compares single videos), the same formulas are applied to one video, skipping sub-metrics and dimensions that do not apply to its script.

\paragraph{Sensitivity to the dimension weights.}
Over 20{,}000 weight vectors drawn uniformly from the simplex over the four dimensions, the base-set ranking agrees with the Bradley--Terry ranking fitted to the pairwise overall preferences (Appendix~\ref{app:human_correlation}) with a median SRCC of 0.83 (95\% range 0.53--0.90). Equal weights reach 0.87, higher than 83\% of the draws and equal to the weights fitted to the pairwise judgments (0.10, 0.41, 0.41, and 0.08 for VQ, DQ, SA, and CSC). With every weight between 0.15 and 0.35, the ranking agrees with the equal-weight ranking at SRCC $\geq$0.97, and the top system is Seedance~2.0 or MiniMax~H3 under 98\% of all draws. Removing one dimension gives SRCC 0.81 without VQ, 0.83 without DQ, 0.87 without SA, and 0.88 without CSC: although Visual Quality alone is uncorrelated with the preference ranking (SRCC 0.00), removing it lowers the agreement the most, so all four dimensions are kept with equal weight.

\section{Metric Visualizations}
\label{app:metric_vis}
For every sub-metric, we show a pair of videos generated from the same script, (a) one with a higher and (b) one with a lower score, on the 0--100 scale of the main tables.

\subsection{Visual Quality}
\label{app:vis_vq}
\begin{figure}[!htbp]
  \centering
  \includegraphics[width=\linewidth]{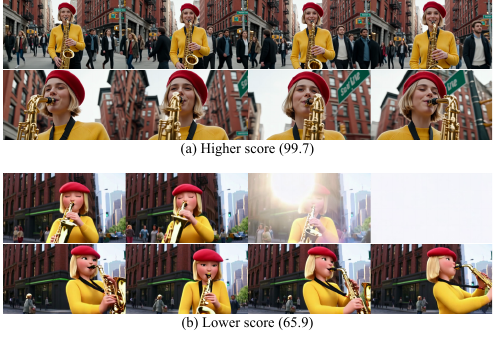}
  \caption{Examples for Aesthetic Score (AS).}
  \label{fig:vis_as}
\end{figure}

\emph{Aesthetic Score} (Figure~\ref{fig:vis_as}). The higher-scoring video keeps a coherent composition and natural lighting in both shots. In the lower-scoring video, the scene exhibits unnaturally high color saturation throughout, and the subject takes on a plastic, cartoon-like appearance, which degrades the overall aesthetic impression.

\begin{figure}[!htbp]
  \centering
  \includegraphics[width=\linewidth]{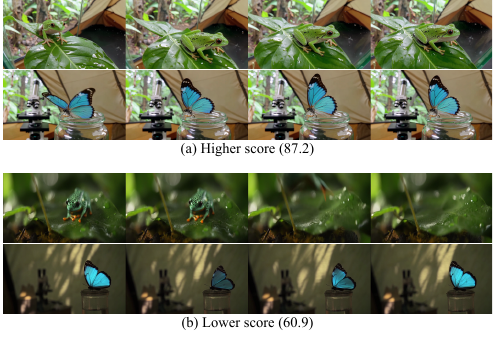}
  \caption{Examples for Perceptual Score (PS).}
  \label{fig:vis_ps}
\end{figure}

\emph{Perceptual Score} (Figure~\ref{fig:vis_ps}). The higher-scoring video renders the frog and the butterfly with sharp textures and clear focus. The lower-scoring video of the same script is dark and out of focus, and fine details such as the skin texture and the wing pattern are lost.

\subsection{Dynamic Quality}
\label{app:vis_dq}
\begin{figure}[!htbp]
  \centering
  \includegraphics[width=\linewidth]{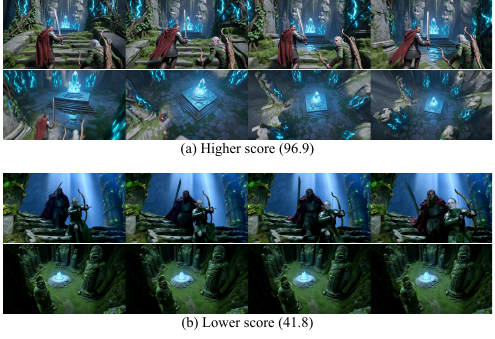}
  \caption{Examples for Dynamic Degree (DD).}
  \label{fig:vis_dd}
\end{figure}

\emph{Dynamic Degree} (Figure~\ref{fig:vis_dd}). In the higher-scoring video, the characters move through the scene in the first shot and the camera sweeps over the altar in the second. In the lower-scoring video, both shots are nearly static: the characters hold their poses and the camera does not move, so DD stays low although each shot is visually clean.

\begin{figure}[!htbp]
  \centering
  \includegraphics[width=\linewidth]{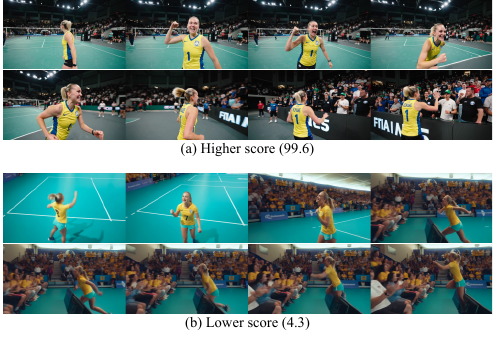}
  \caption{Examples for Physical Plausibility (PP).}
  \label{fig:vis_pp}
\end{figure}

\emph{Physical Plausibility} (Figure~\ref{fig:vis_pp}). The higher-scoring video shows a volleyball player celebrating with continuous and physically coherent body motion. In the lower-scoring video, the player passes through an obstacle while trying to step over it, a body--object interpenetration that the failure checklist flags under object continuity.

\subsection{Semantic Alignment}
\label{app:vis_sa}
\begin{figure}[!htbp]
  \centering
  \includegraphics[width=\linewidth]{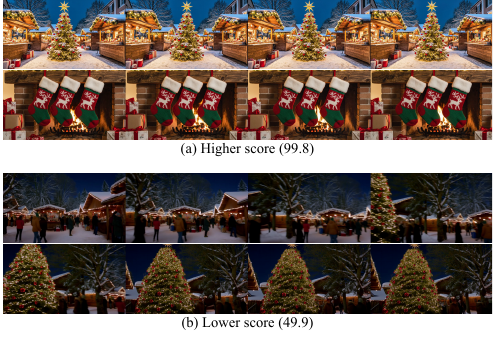}
  \caption{Examples for VQA-Score (VQA).}
  \label{fig:vis_vqa}
\end{figure}

\emph{VQA-Score} (Figure~\ref{fig:vis_vqa}). The script specifies a Christmas tree in a square for the first shot, followed by three Christmas stockings beside a fireplace in the second shot. The higher-scoring video renders both scenes with the requested elements in the intended sequence. In contrast, the lower-scoring video only generates the content of the first shot and completely omits any elements of the fireplace and stockings, causing the VQA questions tied to the second shot to fail and yielding a significantly lower score.

\begin{figure}[!htbp]
  \centering
  \includegraphics[width=\linewidth]{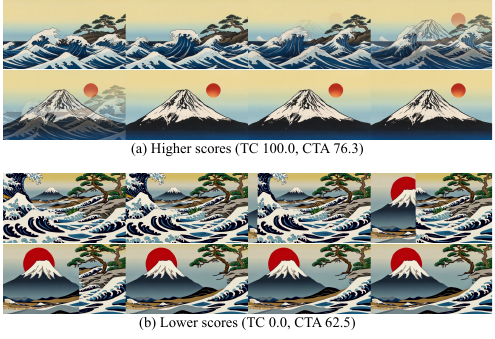}
  \caption{Examples for Transition Control and Cut Timing Accuracy (TC and CTA).}
  \label{fig:vis_tc_cta}
\end{figure}

\emph{Transition Control and Cut Timing Accuracy} (Figure~\ref{fig:vis_tc_cta}). The script requests a dissolve between the two shots. The higher-scoring video blends the waves into the mountain as a dissolve close to the requested time. The lower-scoring video changes shots with a vertical boundary sweeping across the frame, which is detected as a wipe; the transition type is wrong (TC 0), and its timing also deviates more from the script (CTA 62.5 vs.\ 76.3).

\begin{figure}[!htbp]
  \centering
  \includegraphics[width=\linewidth]{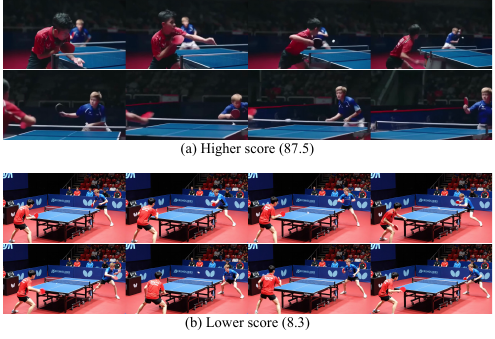}
  \caption{Examples for Camera Control (CC).}
  \label{fig:vis_cc}
\end{figure}

\emph{Camera Control} (Figure~\ref{fig:vis_cc}). The script asks for a medium shot with a dolly followed by a tracking shot. The higher-scoring video follows both requests, whereas the lower-scoring video shows a single fixed long shot of the table, which matches neither the requested scale nor the requested camera movement.

\subsection{Cross-Shot Consistency}
\label{app:vis_csc}
\begin{figure}[!htbp]
  \centering
  \includegraphics[width=\linewidth]{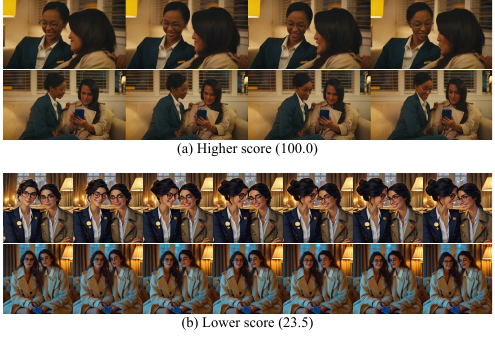}
  \caption{Examples for Cross-Shot Entity Consistency (CSEC).}
  \label{fig:vis_csec}
\end{figure}

\emph{Entity Consistency} (Figure~\ref{fig:vis_csec}). In the higher-scoring video, the two women keep their hairstyles and outfits across the cut. In the lower-scoring video, their clothing and overall appearance change abruptly between the shots, so the same registered entities no longer match.

\begin{figure}[!htbp]
  \centering
  \includegraphics[width=\linewidth]{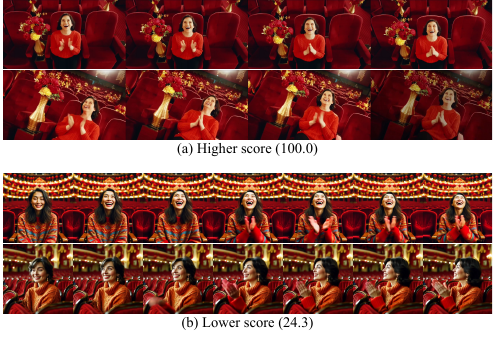}
  \caption{Examples for Cross-Shot Background Consistency (CSBC).}
  \label{fig:vis_csbc}
\end{figure}

\emph{Background Consistency} (Figure~\ref{fig:vis_csbc}). Both videos are set in the same theater. The higher-scoring video keeps the seats and decorations of the hall; in the lower-scoring video, the chairs and the decoration of the hall change between the two shots, although the script places them in the same scene.

\begin{figure}[!htbp]
  \centering
  \includegraphics[width=\linewidth]{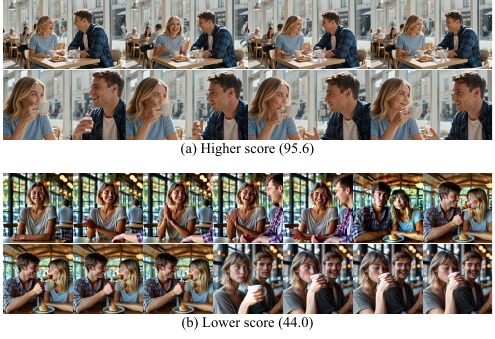}
  \caption{Examples for Cross-Shot Face Consistency (CSFC).}
  \label{fig:vis_csfc}
\end{figure}

\emph{Face Consistency} (Figure~\ref{fig:vis_csfc}). In the higher-scoring video, the woman and the man keep recognizable facial identities from the wide shot to the close-up. In the lower-scoring video, the face of the woman changes across shots, although her clothing stays similar, which the face-level metric isolates from overall entity appearance.

\section{Human Study}
\label{app:human_study}

Two human studies support \name{}. Sub-metric ratings are used to validate every sub-metric (Section~\ref{sec:human}) and to select the Visual and Dynamic Quality methods (Appendix~\ref{app:candidates}); pairwise overall preferences are used to validate the Final Score, both its system ranking and its choices on single pairs of videos (Appendix~\ref{app:human_correlation}). Model identities are hidden in both, and every annotator sees the videos in an independent random order.

\subsection{Sampling}
\label{app:human_sampling}

\paragraph{Sub-metric ratings.}
For each of the four dimensions we annotate 218 videos: 13 base-set scripts rendered by all 14 systems (182 videos) and four 15\,s scripts rendered by the nine systems of the extension (36 videos). Using the same scripts for every system lets us compare systems on identical content, and the scripts are drawn separately for each dimension so that the items of one dimension are not judged on videos an annotator already knows from another. Each video is rated only on the items that apply to its script: the consistency items follow the script's consistency labels, which leaves 152, 146, and 108 videos for entity, background, and face consistency. All applicable items in both annotation rounds are rated by the same five annotators.

\paragraph{Pairwise overall preferences.}
Each pair consists of two videos generated from the same script by two different systems. The 540 distinct pairs comprise 216 pairs from each of the 4\,s and 8\,s tiers, drawn from all 14 systems, and 108 from the 15\,s extension, drawn from its nine systems. Within a tier, system pairings are cycled so that every pair of systems occurs a similar number of times (4 to 9), and scripts are spread evenly: all 300 scripts are used, with at most two pairs per base script and three per 15\,s script. Each system appears in 59 to 90 pairs, fewer for the closed-source systems without 15\,s videos, and on the left side in 49--51\% of its pairs. Three annotators judge the 540 pairs, 30 of which are judged by all three to measure agreement, giving \numPairN{} judgments in total.

\subsection{Annotators and Training}
\label{app:human_annotators}
All annotators are adult volunteers with at least a bachelor's degree and research experience in computer vision. Each was informed of the purpose of the study and of how the ratings would be used and released, and gave consent before participating. No IRB review was carried out, because the study only asks consenting adults to rate synthetic videos, poses minimal risk, and records no personal information beyond anonymous annotator IDs.

Before rating, annotators read the definition and five anchors of every item (Appendix~\ref{app:human_rubrics}) and complete a short trial batch; items with large disagreement are discussed using researcher-reviewed low, medium, and high examples. Scores are absolute: an item is rated against its anchors, not relative to other videos. When the visible evidence is insufficient (\eg, the target face is never visible), annotators select \emph{cannot judge} and give a reason; such items are excluded from the correlations.

\subsection{Annotation Items and Rubrics}
\label{app:human_rubrics}
Annotators rate eleven items on a 1--5 scale, one for each sub-metric. Camera control is rated separately for angle, scale, and movement, each over the whole video. Each item below gives its definition, the five anchors, and the boundary to neighboring items.

\paragraph{Aesthetic quality.}
The composition, color, lighting, and visual style of the video and their overall harmony. Realistic, animated, and painterly styles are judged by the same principles.
\textbf{1}: composition or color is clearly unbalanced; the result looks cluttered and unfinished.
\textbf{2}: a few frames are attractive, but the main frames have clearly mismatched composition, color, or style.
\textbf{3}: composition and color basically work and have some appeal, with clear but local problems.
\textbf{4}: most frames have clear composition and harmonious color and lighting, with a few minor problems.
\textbf{5}: mature visual design throughout, with composition, color, and lighting working together and sustained appeal.
\emph{Boundary}: personal taste for subject or style is not a criterion; sharpness and artifacts belong to technical quality, and content to semantic alignment.

\paragraph{Technical quality.}
Sharpness, detail and texture fidelity, and the effect of generation artifacts (blur, noise, broken structure, texture distortion, flicker) on legibility and stability.
\textbf{1}: most frames are severely blurred, broken, or unstable; the subject is hard to recognize.
\textbf{2}: clear defects recur or persist and seriously hurt the legibility of subjects, structure, or detail.
\textbf{3}: the main content is recognizable, but local blur, texture distortion, edge flicker, or structural artifacts are noticeable.
\textbf{4}: clear and stable overall, with a few small, brief, or local defects.
\textbf{5}: clear and stable with natural detail and texture; no noticeable technical defect at normal playback.
\emph{Boundary}: motion blur, depth of field, and intended stylization are not defects.

\paragraph{Motion magnitude.}
How much visible motion the video contains, from subjects, the environment, and the camera, considering amplitude, spatial extent, duration, and variation; a single action or camera move can suffice.
\textbf{1}: nearly static. \textbf{2}: limited amplitude, extent, or duration. \textbf{3}: clear motion, but limited in amplitude, extent, or duration. \textbf{4}: substantial motion during most of the video. \textbf{5}: sustained, pronounced motion or rich variation throughout.
\emph{Boundary}: naturalness and physical plausibility are rated separately; compression flicker, edge jitter, structural collapse, and the cut or transition itself are not motion.

\paragraph{Physical plausibility.}
Whether motion, deformation, contact, and interaction are physically credible and causally coherent within the video's setting.
\textbf{1}: persistent severe contradictions in the main action, such as interpenetration, unsupported drifting, collapsing bodies, or broken cause and effect.
\textbf{2}: several clearly implausible motions or interactions recur.
\textbf{3}: mostly understandable, with at least one clear motion, contact, or deformation error.
\textbf{4}: natural overall, with slight or brief flaws.
\textbf{5}: all observable actions and interactions are coherent and credible.
\emph{Boundary}: magic or levitation established by the setting is not an error; a static video earns no bonus, and \emph{cannot judge} is used when there is too little motion to assess.

\paragraph{Semantic content (VQA).}
Whether the scenes, subjects, attributes, actions, and relations of the script are present and correct in every segment, combined into one score.
\textbf{1}: unrelated content, or most core scenes, subjects, and actions are missing or wrong.
\textbf{2}: part of the content appears, but several important subjects, attributes, actions, or relations are missing or wrong.
\textbf{3}: subjects and scenes roughly correct and the story understandable, with clear important omissions.
\textbf{4}: core content essentially complete and correct, with minor omissions.
\textbf{5}: core content complete and accurate in every segment.
\emph{Boundary}: camera, transitions, and exact timing are rated separately; scene changes and action order that belong to the story are included.

\paragraph{Transition control.}
Whether each transition between corresponding shots has the requested type (hard cut, dissolve, fade, wipe) and is clearly and completely executed.
\textbf{1}: no requested transition is realized, or all have the wrong type.
\textbf{2}: only a few are recognizable; most are missing, wrong, or badly incomplete.
\textbf{3}: some have the correct type, with clear errors, omissions, or ambiguity.
\textbf{4}: most are correct and clear, with one slightly incomplete or unclear execution.
\textbf{5}: all are correct and clean, with no extra transition that breaks the shot structure.
\emph{Boundary}: timing is rated by cut timing; continuous camera motion or morphing without a shot change is not a transition.

\paragraph{Cut timing.}
How well the actual shot boundaries match the requested timeline: the time of each hard cut, the start and end of each gradual transition, missing or extra boundaries, and the allocation of shot durations.
\textbf{1}: no boundary corresponds, or the shot count, order, and timing are badly mismatched.
\textbf{2}: only a few boundaries correspond, or most are far too early or late.
\textbf{3}: the main boundaries correspond in order, with at least one clear timing error, missing, or extra boundary.
\textbf{4}: correct count and order, most boundaries close to target, with small deviations.
\textbf{5}: all boundaries correct in count and order and nearly coincident with the target.
\emph{Boundary}: a wrong transition type alone is not penalized here.

\paragraph{Camera control (angle, scale, movement).}
Each aspect is rated over all shots of the video. Angle is judged from the camera height relative to the subject and the frame tilt; scale from the framing of the subject; movement from changes of viewpoint, perspective, and framing (a dolly changes perspective, a zoom mainly changes magnification; subject motion is not camera motion).
\textbf{1}: the requirement is essentially unmet in every shot. \textbf{2}: met in only a few shots. \textbf{3}: met in some shots, with clear or repeated deviations. \textbf{4}: met in most shots, with small or brief deviations. \textbf{5}: met accurately in every shot.

\paragraph{Entity consistency.}
Whether the same person, animal, or object keeps its identity and key appearance across shots: shape, body proportions, colors, materials, clothing, patterns, and key parts.
\textbf{1}: a matched entity turns into a different identity or category, or its structure is completely incoherent.
\textbf{2}: several large appearance changes.
\textbf{3}: identity kept, but at least one key attribute clearly changes.
\textbf{4}: identity and key attributes stable, with a few small local changes.
\textbf{5}: identity, structure, and key attributes stable in every shot.
\emph{Boundary}: differences explained by viewpoint, pose, occlusion, lighting, or a requested state change are not penalized; facial identity belongs to face consistency.

\paragraph{Background consistency.}
Whether the same environment keeps its spatial structure, layout, and fixed features across shots: scene identity, fixed buildings and furniture, layout, materials and textures, and landmarks.
\textbf{1}: shots meant to share an environment show a different place, or the fixed structure is contradictory.
\textbf{2}: main layout, buildings, or fixed objects change substantially in several places.
\textbf{3}: scene identity kept, but at least one important feature clearly drifts.
\textbf{4}: matching features essentially stable, with small local differences.
\textbf{5}: fixed structure and layout stable; differences are explained by viewpoint or framing.
\emph{Boundary}: scenes the script intends to be different are not compared; viewpoint, scale, and lighting differences are not penalized by themselves.

\paragraph{Face consistency.}
Whether the same person keeps the same facial identity across shots: face shape, facial features, their proportions and spacing.
\textbf{1}: the facial identity clearly changes. \textbf{2}: face shape or several features change substantially. \textbf{3}: identity recognizable, with at least one clear drift in a feature or proportion. \textbf{4}: stable, with slight local changes. \textbf{5}: identity and key features stable; differences are explained by expression, viewpoint, or lighting.
\emph{Boundary}: expression, head pose, perspective, lighting, and makeup are not identity changes; \emph{cannot judge} is used when faces are too small, turned away, or occluded.

\paragraph{Overall preference.}
For a pair of videos generated from the same script, annotators choose which video is better as a whole (A better, about the same, B better). No weighting of visual quality, motion, script adherence, or consistency is prescribed, so the judgment reflects the annotator's own overall impression of the video as a realization of the script. The two videos are shown side by side with a one-sentence summary of each shot and the requested transitions; the summaries are generated from the script by Qwen3.8-27B, and the full script can be expanded on the same page. Videos are muted, and both videos must be played to the end before an answer can be given.

\subsection{Agreement with Human Ratings}
\label{app:human_correlation}
Ratings are averaged over annotators into a mean opinion score (MOS) per video and item. Table~\ref{tab:human_full} compares each sub-metric with the MOS on the rated videos to which it applies, at the video level (SRCC with a 95\% bootstrap interval, KRCC, and PLCC) and at the model level (SRCC after averaging both per system). The \emph{ICC} column gives the reliability of the MOS as the average-measure ICC(1,5).

\begin{table*}[!t]
  \centering
  \caption{Agreement ($\times$100) between \name{} sub-metrics and human ratings. $^\ast$In-sample; the cross-validated SRCC is \numDDcv{}.}
  \label{tab:human_full}
  \footnotesize
  \setlength{\tabcolsep}{5pt}
  \begin{tabular*}{\linewidth}{@{\extracolsep{\fill}}lccccccc@{}}
    \toprule
    \textbf{Sub-metric} & \textbf{\#Videos} & \textbf{SRCC} & \textbf{95\% CI} & \textbf{KRCC} & \textbf{PLCC} & \textbf{Model} & \textbf{ICC} \\
    \midrule
    Aesthetic Score          & 218 & 46.0 & [34, 56] & 32.9 & 49.2 & 50.8 & 0.68 \\
    Perceptual Score         & 218 & 39.7 & [26, 51] & 28.5 & 46.4 & 30.5 & 0.79 \\
    Dynamic Degree           & 218 & 67.7$^\ast$ & [59, 74] & 49.9 & 67.5 & 71.9 & 0.87 \\
    Physical Plausibility    & 218 & 30.5 & [19, 42] & 20.8 & 28.1 & 66.2 & 0.79 \\
    VQA-Score                & 218 & 64.6 & [55, 72] & 47.7 & 66.5 & 85.5 & 0.87 \\
    Transition Control       & 218 & 89.6 & [87, 92] & 79.6 & 92.0 & 95.3 & 0.97 \\
    Cut Timing Accuracy      & 218 & 81.7 & [74, 88] & 67.2 & 78.5 & 97.7 & 0.93 \\
    Camera Control           & 218 & 66.1 & [57, 73] & 48.7 & 68.7 & 87.3 & 0.91 \\
    Entity Consistency       & 152 & 46.3 & [33, 59] & 34.2 & 60.2 & 61.0 & 0.81 \\
    Background Consistency   & 146 & 39.4 & [24, 54] & 28.9 & 46.4 & 84.7 & 0.76 \\
    Face Consistency         & 108 & 63.2 & [49, 74] & 47.8 & 48.4 & 68.1 & 0.87 \\
 
    \bottomrule
  \end{tabular*}
\end{table*}

The ICC is lowest for aesthetics and background consistency (0.68 and 0.76), where judgments depend more on personal taste; Physical Plausibility is discussed in Appendix~\ref{app:candidates}. For the pairwise preferences, the three-way agreement on the 30 shared pairs is 63.3\% (Fleiss' $\kappa$ 0.36), and one annotator predicts another in 77.0\% of the 74 non-tie annotator pairs.

\paragraph{System ranking from pairwise preferences.}
We fit a Bradley--Terry model over the 14 systems, $P(i\succ j)=\sigma(\theta_i-\theta_j)$, to all \numPairN{} judgments, counting a tie as half a win for each side and estimating $\theta$ by minorization--maximization with a pseudo-count of 0.5 wins per system; each system enters \numPairPerSysLo{} to \numPairPerSysHi{} judgments. The ranking by $\theta$ agrees with the ranking by base-set Final Score with SRCC \numFinalBT{} (Kendall's $\tau$ \numFinalBTk{}; 95\% bootstrap interval [0.75, 0.90] over 1{,}000 resamples of the judgments), and the Final Score over all tiers gives the same SRCC.
Since such a correlation could come only from the gap between strong and weak systems, we repeat it on the nine leading systems (base-set Final Score 66.8--79.6): the SRCC is 0.97 ($\tau$ 0.89; [0.80, 0.98]) with the strengths fitted on all judgments, and 0.87 ($\tau$ 0.72) when Bradley--Terry is refitted on only the 241 judgments among these systems.

\paragraph{Pairwise accuracy of the Final Score.}
On single pairs of videos, the Final Score picks the preferred video in 67.1\% of the 560 non-tie judgments, compared with 50.0--63.6\% for any single dimension and 77.0\% for one annotator predicting another.

\subsection{Candidate Methods and Selection}
\label{app:candidates}
For Visual and Dynamic Quality, where off-the-shelf scorers are most numerous, we chose each sub-metric among several candidate methods by its agreement with the human ratings, with interpretability and determinism as secondary criteria. All candidates were run with official weights and default settings on the same detected shots and scored on the same 218 human-rated videos. Table~\ref{tab:candidates} lists the candidates.

\begin{table}[!htbp]
  \centering
  \caption{Candidate methods for the Visual and Dynamic Quality sub-metrics: SRCC ($\times$100) with human ratings. Selected methods are in bold. $^\dagger$Cross-validated.}
  \label{tab:candidates}
  \footnotesize
  \setlength{\tabcolsep}{3pt}
  \begin{tabular*}{\linewidth}{@{\extracolsep{\fill}}llcc@{}}
    \toprule
     & \textbf{Candidate} & \textbf{SRCC} & \textbf{Model} \\
    \midrule
      \multirow{4}{*}{AS} & \textbf{Q-Align~\citep{wu2023qalign}} & 46.0 & 50.8 \\
       & Aesthetic-Predictor V2.5~\citep{aesthetic_predictor_v25} & 33.1 & 27.5 \\
       & DOVER++ aesthetic branch~\citep{wu2023dover} & 20.6 & 38.9 \\
       & LAION aesthetic predictor~\citep{laion_aesthetic_predictor} & 9.8 & 23.5 \\
      \midrule
      \multirow{4}{*}{PS} & \textbf{DeQA-Score~\citep{deqa_score}} & 39.7 & 30.5 \\
       & FAST-VQA~\citep{wu2022fastvqa} & 32.2 & 31.4 \\
       & DOVER++ technical branch~\citep{wu2023dover} & 29.1 & 44.2 \\
       & MUSIQ~\citep{ke2021musiq} & 19.2 & 5.5 \\
 
    \midrule
      \multirow{3}{*}{DD} & \textbf{DEVIL components, fitted weights}$^\dagger$ & 60.7 & 69.4 \\
       & Continuous RAFT flow~\citep{zheng2025locot2v} & 63.1 & 46.8 \\
       & DEVIL official fusion~\citep{liao2024devil} & 55.3 & 23.1 \\
      \midrule
      \multirow{2}{*}{PP} & \textbf{Failure checklist, strict top anchor} & 30.5 & 66.2 \\
       & VideoPhy-2-AutoEval & $-$2.5 & $-$13.0 \\
 
    \bottomrule
  \end{tabular*}
\end{table}

Two patterns guided the choices. For visual quality, predictors trained on natural images or user-generated videos (LAION aesthetic, MUSIQ, DOVER++) transfer poorly to generated videos, whereas the LMM-based Q-Align and DeQA-Score agree best within their groups. For Dynamic Degree, we weigh video-level and system-level agreement together: continuous RAFT flow is slightly higher at the video level (63.1 vs.\ \numDDcv{}), but the DEVIL components with fitted weights rank systems far more like the annotators (\numDDcvModel{} vs.\ 46.8). For Physical Plausibility, we use a Qwen3.8 failure checklist read as an expected score (Appendix~\ref{app:dynamic_details}). PP remains the weakest sub-metric at the video level: a general-purpose VLM prompted with a failure checklist overlooks many of the subtle failures that annotators penalize, and two judges agree with each other only weakly on it (Appendix~\ref{app:judge_robustness}). The open-source physics expert VideoPhy-2-AutoEval is uncorrelated with our annotators, likely because multi-shot cinematic videos lie outside its training distribution. At the system level, however, the selected checklist separates systems well (model-level SRCC 66.2), so we recommend reading PP at the system level and together with the other sub-metrics.

\subsection{Robustness to the VLM Judge}
\label{app:judge_robustness}

\begin{table}[!htbp]
  \centering
  \caption{Agreement between the Qwen3.8 and Doubao-Seed-2.1-lite judges (SRCC).}
  \label{tab:judge_robustness}
  \footnotesize
  \begin{tabular*}{\linewidth}{@{\extracolsep{\fill}}lcc@{}}
    \toprule
    \textbf{Sub-metric} & \textbf{Video SRCC} & \textbf{Model SRCC} \\
    \midrule
    PP & 0.26 & 0.82 \\
    VQA & 0.88 & 0.98 \\
    CSEC & 0.46 & 0.75 \\
    CSBC & 0.47 & 0.85 \\
 
    \bottomrule
  \end{tabular*}
\end{table}

To test how much the conclusions depend on Qwen3.8, we re-score the four VLM-based sub-metrics on all videos with Doubao-Seed-2.1-lite, a model from a different family, using the same prompts, the same videos, and the same score reading from token probabilities; because its API accepts at most 5\,fps, PP, CSEC, and CSBC are sampled at 5 instead of 10\,fps. Table~\ref{tab:judge_robustness} reports the agreement. The two judges agree closely on VQA and, at the system level, on entity and background consistency, and least on Physical Plausibility, which is also the sub-metric hardest to align with humans. In the base-set ranking by Final Score, only Veo3.1 and Sora2~Pro, 0.04 apart, swap places.

\section{Additional Quantitative Results}
\label{app:additional_results}

This section first gives the overall scores and the complete per-tier results, followed by the breakdowns behind the findings of Section~\ref{sec:insights}.

\subsection{Overall Scores and Ranking Stability}
\label{app:overall}

\begin{figure*}[!t]
  \centering
  \includegraphics[width=0.92\linewidth]{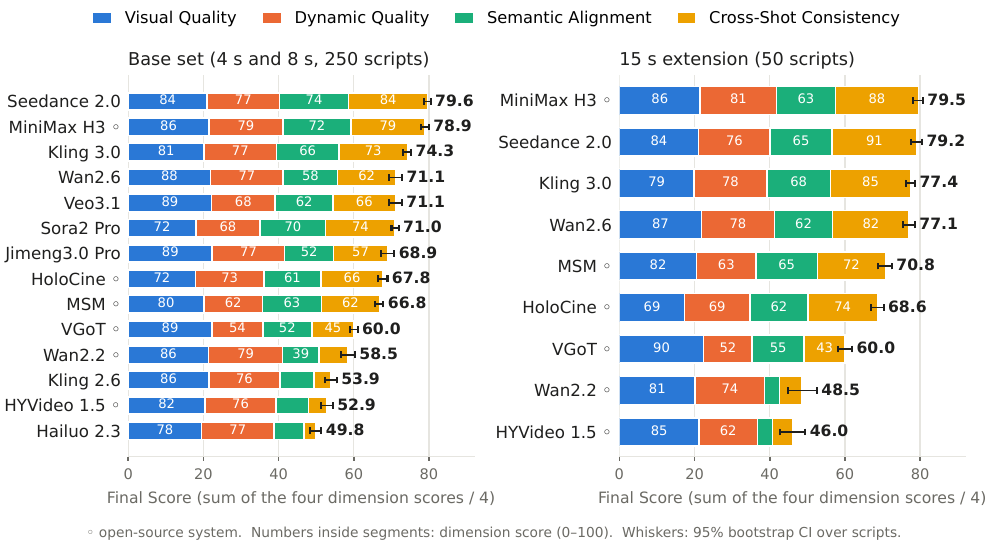}
  \caption{Final Score of each system, decomposed into its four dimension scores (each divided by four).}
  \label{fig:overall_scores}
\end{figure*}

Figure~\ref{fig:overall_scores} decomposes the Final Score of every system into its four dimension scores. Visual and Dynamic Quality contribute a similar amount for almost all systems, so the ranking is decided by Semantic Alignment and Cross-Shot Consistency, the two dimensions specific to multi-shot generation.

To see how much the ranking depends on the particular scripts, we resample the scripts with replacement 2{,}000 times (Table~\ref{tab:final_ci}). On the base set the 95\% interval of the Final Score is $\pm$0.8 to $\pm$1.8 points, and the ranking separates into stable groups: Seedance~2.0 and MiniMax~H3 are tied at the top, Kling~3.0 is third in every resample, and the five lowest-ranked systems never mix with the upper nine. Rankings computed on a random half of the scripts agree with the full ranking with a mean Kendall's $\tau$ of \numStabBase{} on the base set and \numStabFifteen{} on the 15\,s extension, whose intervals are wider because it has 50 scripts.

\begin{table*}[!t]
  \centering
  \caption{Final Score and rank with 95\% bootstrap intervals over scripts.}
  \label{tab:final_ci}
  \footnotesize
  \setlength{\tabcolsep}{4pt}
  \begin{tabular*}{\linewidth}{@{\extracolsep{\fill}}lcccccccc@{}}
    \toprule
     & \multicolumn{4}{c}{\textbf{Base set}} & \multicolumn{4}{c}{\textbf{15\,s extension}} \\
    \cmidrule(lr){2-5} \cmidrule(lr){6-9}
    \textbf{Model} & Final & 95\% CI & Rank & Rank CI & Final & 95\% CI & Rank & Rank CI \\
    \midrule
    Seedance~2.0     & 79.6 & [78.8, 80.4] & 1 & 1--2 & 79.2 & [77.7, 80.5] & 2 & 1--2 \\
    MiniMax~H3       & 78.9 & [77.8, 79.9] & 2 & 1--2 & 79.5 & [78.1, 80.8] & 1 & 1--2 \\
    Kling~3.0        & 74.3 & [73.1, 75.3] & 3 & 3--3 & 77.4 & [76.1, 78.7] & 3 & 3--4 \\
    Wan2.6           & 71.1 & [69.4, 72.9] & 4 & 4--7 & 77.1 & [75.5, 78.6] & 4 & 3--4 \\
    Veo3.1           & 71.1 & [69.3, 72.8] & 5 & 4--7 & -- & -- & -- & -- \\
    Sora2~Pro        & 71.0 & [70.0, 71.9] & 6 & 4--6 & -- & -- & -- & -- \\
    Jimeng3.0~Pro    & 68.9 & [67.1, 70.8] & 7 & 5--8 & -- & -- & -- & -- \\
    HoloCine         & 67.8 & [66.4, 69.0] & 8 & 7--9 & 68.6 & [66.9, 70.3] & 6 & 6--6 \\
    MSM              & 66.8 & [65.7, 67.9] & 9 & 8--9 & 70.8 & [68.8, 72.5] & 5 & 5--5 \\
    VGoT             & 60.0 & [58.9, 61.0] & 10 & 10--11 & 60.0 & [58.2, 61.9] & 7 & 7--7 \\
    Wan2.2           & 58.5 & [56.7, 60.3] & 11 & 10--11 & 48.5 & [44.8, 52.6] & 8 & 8--9 \\
    Kling~2.6        & 53.9 & [52.4, 55.5] & 12 & 12--13 & -- & -- & -- & -- \\
    HYVideo~1.5      & 52.9 & [51.4, 54.4] & 13 & 12--13 & 46.0 & [42.7, 49.5] & 9 & 8--9 \\
    Hailuo~2.3       & 49.8 & [48.4, 51.4] & 14 & 14--14 & -- & -- & -- & -- \\
 
    \bottomrule
  \end{tabular*}
\end{table*}

\subsection{Full Results per Tier}
\label{app:full_results}

Tables~\ref{tab:full_4s}--\ref{tab:full_15s} report all sub-metric, dimension, and Final scores for each tier. Here the cross-shot columns are the scores that enter CSC, \ie, invalid applicable videos count as zero.

\begin{table*}[!t]
  \centering
  \caption{Full results on the 4\,s tier (0--100). Rank colors and $^\dagger$ as in Table~\ref{tab:main_results}.}
  \label{tab:full_4s}
  \scriptsize
  \setlength{\tabcolsep}{3.2pt}
  \begin{tabular*}{\linewidth}{@{\extracolsep{\fill}}l*{18}{c}@{}}
    \toprule
    \multirow{2}{*}{\textbf{Model}} & \multicolumn{2}{c}{\textbf{VQ}} & \multicolumn{2}{c}{\textbf{DQ}} & \multicolumn{6}{c}{\textbf{SA}} & \multicolumn{3}{c}{\textbf{CSC}} & \multicolumn{5}{c}{\textbf{Dimension and Final}} \\
    \cmidrule(lr){2-3} \cmidrule(lr){4-5} \cmidrule(lr){6-11} \cmidrule(lr){12-14} \cmidrule(lr){15-19}
     & AS & PS & DD & PP & VQA & TC-V & TC-A & CTA & CC-V & CC & CSEC & CSBC & CSFC & VQ & DQ & SA & CSC & Final \\
    \midrule
    \multicolumn{19}{c}{\cellcolor{categorybg}\textbf{Conventional T2V Models}} \\
    Kling~2.6              & 94.8 & 74.5 & 68.8 & 87.5 & 72.8 & 29.6 & 21.6 & 32.8 & 64.8 & 28.8 & 19.2 & 5.8 & 4.9 & 84.7 & 78.1 & 35.2 & 10.0 & 52.0 \\
    Hailuo~2.3             & 87.1 & 65.2 & 72.6 & 84.1 & 67.6 & 16.0 & 35.0 & 26.6 & 58.0 & 26.7 & 10.7 & 5.4 & 1.9 & 76.2 & 78.4 & 31.6 & 6.0 & 48.0 \\
    Wan2.2$^\dagger$       & 94.7 & 76.3 & \best{rankA}{76.8} & 87.3 & 75.6 & 46.4 & 43.1 & 39.2 & 72.8 & 30.7 & 30.5 & 23.9 & 12.4 & 85.5 & \best{rankA}{82.1} & 41.4 & 22.2 & 57.8 \\
    HYVideo~1.5$^\dagger$  & 88.0 & 72.0 & 71.5 & 81.5 & 71.6 & 36.0 & 22.2 & 33.3 & 67.2 & 29.8 & 23.3 & 11.4 & 13.9 & 80.0 & 76.5 & 35.7 & 16.2 & 52.1 \\
    \midrule
    \multicolumn{19}{c}{\cellcolor{categorybg}\textbf{Native Multi-shot T2V Models}} \\
    Sora2~Pro              & 78.1 & 62.8 & 53.1 & 85.1 & 83.2 & \cell{rankB}{98.4} & 79.7 & \cell{rankD}{80.6} & \cell{rankB}{98.8} & 41.6 & \cell{rankB}{90.6} & 76.7 & \cell{rankD}{51.3} & 70.4 & 69.1 & \cell{rankD}{70.9} & \cell{rankD}{72.9} & 70.9 \\
    Veo3.1                 & \cell{rankD}{96.5} & \cell{rankD}{80.6} & 55.0 & 81.2 & 84.2 & 81.6 & \cell{rankD}{82.4} & 50.9 & 88.0 & 31.5 & 67.2 & 61.2 & 36.6 & \cell{rankD}{88.5} & 68.1 & 58.4 & 55.0 & 67.5 \\
    Wan2.6                 & 95.7 & 78.8 & \cell{rankB}{75.5} & 81.5 & 83.7 & 75.2 & 53.2 & 58.1 & 86.4 & 37.6 & 69.0 & 53.2 & 33.6 & 87.2 & \cell{rankD}{78.5} & 54.9 & 51.9 & 68.1 \\
    Jimeng3.0~Pro          & \best{rankA}{96.8} & \cell{rankB}{80.7} & 69.0 & \cell{rankB}{87.8} & 75.5 & 80.8 & 55.4 & 60.8 & 89.6 & 44.7 & 67.8 & 54.8 & 47.7 & \cell{rankB}{88.7} & 78.4 & 56.5 & 56.8 & 70.1 \\
    Seedance~2.0           & 92.4 & 73.0 & 64.0 & \best{rankA}{91.3} & \best{rankA}{88.9} & 96.0 & \cell{rankB}{85.0} & 72.9 & 98.0 & \best{rankA}{55.0} & \best{rankA}{95.1} & \best{rankA}{82.0} & \best{rankA}{73.1} & 82.7 & 77.6 & \best{rankA}{74.6} & \best{rankA}{83.4} & \best{rankA}{79.6} \\
    Kling~3.0              & 88.3 & 71.5 & 69.0 & 84.8 & \cell{rankB}{86.9} & 95.2 & 48.7 & \cell{rankB}{83.0} & 97.2 & \cell{rankD}{51.7} & \cell{rankD}{88.2} & \cell{rankD}{77.0} & 38.8 & 79.9 & 76.9 & 67.0 & 68.0 & \cell{rankD}{72.9} \\
    MiniMax~H3$^\dagger$   & 93.1 & 76.9 & \cell{rankD}{74.1} & \cell{rankD}{87.7} & \cell{rankB}{86.9} & 94.4 & \best{rankA}{92.4} & 59.7 & 96.4 & \cell{rankB}{54.5} & 88.1 & \cell{rankB}{80.9} & \cell{rankB}{58.3} & 85.0 & \cell{rankB}{80.9} & \cell{rankB}{72.1} & \cell{rankB}{75.8} & \cell{rankB}{78.4} \\
    \midrule
    \multicolumn{19}{c}{\cellcolor{categorybg}\textbf{Multi-shot Specialists}} \\
    HoloCine$^\dagger$     & 78.8 & 62.2 & 60.8 & 87.2 & 83.3 & 90.4 & 53.1 & 74.5 & 95.2 & 40.2 & 81.8 & 60.2 & 43.7 & 70.5 & 74.0 & 61.5 & 61.9 & 67.0 \\
    MSM$^\dagger$          & 89.1 & 69.3 & 45.2 & 78.7 & 81.3 & \best{rankA}{99.2} & 46.0 & \best{rankA}{88.4} & \best{rankA}{99.6} & 37.9 & 82.7 & 69.4 & 23.7 & 79.2 & 61.9 & 63.3 & 58.6 & 65.8 \\
    VGoT$^\dagger$         & \cell{rankB}{96.6} & \best{rankA}{81.4} & 45.8 & 67.1 & 70.0 & \cell{rankB}{98.4} & 32.5 & 69.5 & \cell{rankB}{98.8} & 36.4 & 49.9 & 38.8 & 46.0 & \best{rankA}{89.0} & 56.5 & 52.0 & 44.9 & 60.6 \\
 
    \bottomrule
  \end{tabular*}
\end{table*}

\begin{table*}[!t]
  \centering
  \caption{Full results on the 8\,s tier (0--100).}
  \label{tab:full_8s}
  \scriptsize
  \setlength{\tabcolsep}{3.2pt}
  \begin{tabular*}{\linewidth}{@{\extracolsep{\fill}}l*{18}{c}@{}}
    \toprule
    \multirow{2}{*}{\textbf{Model}} & \multicolumn{2}{c}{\textbf{VQ}} & \multicolumn{2}{c}{\textbf{DQ}} & \multicolumn{6}{c}{\textbf{SA}} & \multicolumn{3}{c}{\textbf{CSC}} & \multicolumn{5}{c}{\textbf{Dimension and Final}} \\
    \cmidrule(lr){2-3} \cmidrule(lr){4-5} \cmidrule(lr){6-11} \cmidrule(lr){12-14} \cmidrule(lr){15-19}
     & AS & PS & DD & PP & VQA & TC-V & TC-A & CTA & CC-V & CC & CSEC & CSBC & CSFC & VQ & DQ & SA & CSC & Final \\
    \midrule
    \multicolumn{19}{c}{\cellcolor{categorybg}\textbf{Conventional T2V Models}} \\
    Kling~2.6              & 96.5 & 77.1 & 61.8 & 85.1 & 66.9 & 39.9 & 42.4 & 30.7 & 63.4 & 31.1 & 30.1 & 31.3 & 16.4 & 86.8 & 73.4 & 36.4 & 25.9 & 55.6 \\
    Hailuo~2.3             & 90.2 & 68.8 & 61.2 & \cell{rankB}{91.3} & 62.5 & 25.3 & 66.7 & 21.9 & 54.1 & 26.8 & 19.7 & 15.7 & 17.1 & 79.5 & 76.2 & 32.0 & 17.5 & 51.3 \\
    Wan2.2$^\dagger$       & 94.0 & 77.0 & 60.8 & 89.6 & 65.4 & 48.3 & 32.4 & 31.1 & 69.8 & 32.1 & 48.8 & 34.1 & 32.7 & 85.5 & 75.2 & 36.0 & 38.5 & 58.8 \\
    HYVideo~1.5$^\dagger$  & 92.3 & 74.5 & \cell{rankD}{64.9} & 84.2 & 64.0 & 33.5 & 32.2 & 27.9 & 60.3 & 29.9 & 27.7 & 27.3 & 15.6 & 83.4 & 74.6 & 33.1 & 23.6 & 53.7 \\
    \midrule
    \multicolumn{19}{c}{\cellcolor{categorybg}\textbf{Native Multi-shot T2V Models}} \\
    Sora2~Pro              & 81.6 & 64.8 & 50.8 & 84.5 & 82.8 & 98.0 & 67.7 & 74.5 & \cell{rankD}{98.8} & 50.3 & 90.6 & 85.5 & 50.3 & 73.2 & 67.7 & \cell{rankD}{68.5} & 75.4 & 71.2 \\
    Veo3.1                 & 96.5 & \cell{rankD}{80.5} & 53.1 & 81.4 & 85.3 & 93.1 & \cell{rankB}{83.5} & 51.0 & 95.1 & 47.5 & 90.0 & \cell{rankD}{86.1} & \cell{rankD}{52.7} & \cell{rankD}{88.5} & 67.3 & 65.4 & 76.3 & 74.4 \\
    Wan2.6                 & \cell{rankD}{96.8} & 79.4 & \best{rankA}{66.1} & 86.0 & 80.0 & 94.9 & 50.6 & 65.4 & 97.1 & 49.5 & 89.0 & 78.5 & 44.8 & 88.1 & 76.1 & 60.7 & 70.7 & 73.9 \\
    Jimeng3.0~Pro          & \best{rankA}{98.3} & \best{rankA}{82.0} & 63.4 & 87.9 & 66.8 & 76.8 & 38.9 & 48.3 & 85.6 & 46.5 & 62.3 & 69.5 & 42.7 & \best{rankA}{90.2} & 75.6 & 47.9 & 58.2 & 68.0 \\
    Seedance~2.0           & 94.6 & 75.6 & 59.5 & \best{rankA}{93.2} & \best{rankA}{88.3} & \cell{rankD}{98.1} & \cell{rankD}{75.6} & 67.0 & 98.7 & \cell{rankB}{61.1} & \cell{rankB}{93.2} & \cell{rankB}{87.1} & \best{rankA}{73.8} & 85.1 & \cell{rankD}{76.3} & \best{rankA}{72.7} & \best{rankA}{84.7} & \best{rankA}{79.7} \\
    Kling~3.0              & 90.2 & 72.9 & 62.6 & \cell{rankD}{91.1} & \cell{rankB}{87.3} & \best{rankA}{99.6} & 31.1 & \cell{rankB}{81.9} & \best{rankA}{99.7} & \best{rankA}{63.8} & \cell{rankD}{91.5} & \best{rankA}{90.6} & 50.7 & 81.6 & \cell{rankB}{76.9} & 66.0 & \cell{rankD}{77.6} & \cell{rankD}{75.5} \\
    MiniMax~H3$^\dagger$   & 95.1 & 77.7 & \cell{rankB}{65.5} & 88.8 & \cell{rankD}{86.3} & 96.0 & \best{rankA}{87.7} & 59.9 & 97.6 & \cell{rankD}{57.9} & \best{rankA}{93.3} & 84.9 & \cell{rankB}{66.4} & 86.4 & \best{rankA}{77.1} & \cell{rankB}{72.1} & \cell{rankB}{81.5} & \cell{rankB}{79.3} \\
    \midrule
    \multicolumn{19}{c}{\cellcolor{categorybg}\textbf{Multi-shot Specialists}} \\
    HoloCine$^\dagger$     & 81.4 & 63.7 & 54.3 & 88.2 & 81.6 & 96.8 & 34.7 & \cell{rankD}{75.3} & 98.1 & 51.4 & 90.9 & 77.5 & 42.1 & 72.5 & 71.2 & 60.5 & 70.2 & 68.6 \\
    MSM$^\dagger$          & 91.2 & 72.2 & 41.3 & 84.6 & 80.0 & \best{rankA}{99.6} & 31.1 & \best{rankA}{85.5} & \cell{rankB}{99.5} & 50.9 & 87.7 & 76.6 & 29.5 & 81.7 & 63.0 & 61.8 & 64.6 & 67.8 \\
    VGoT$^\dagger$         & \cell{rankB}{97.0} & \cell{rankB}{80.9} & 44.7 & 59.0 & 69.5 & 97.2 & 23.3 & 69.1 & 97.7 & 47.8 & 44.6 & 43.9 & 46.6 & \cell{rankB}{89.0} & 51.9 & 52.3 & 45.0 & 59.5 \\
 
    \bottomrule
  \end{tabular*}
\end{table*}

\begin{table*}[!t]
  \centering
  \caption{Full results on the 15\,s extension (0--100).}
  \label{tab:full_15s}
  \scriptsize
  \setlength{\tabcolsep}{3.2pt}
  \begin{tabular*}{\linewidth}{@{\extracolsep{\fill}}l*{18}{c}@{}}
    \toprule
    \multirow{2}{*}{\textbf{Model}} & \multicolumn{2}{c}{\textbf{VQ}} & \multicolumn{2}{c}{\textbf{DQ}} & \multicolumn{6}{c}{\textbf{SA}} & \multicolumn{3}{c}{\textbf{CSC}} & \multicolumn{5}{c}{\textbf{Dimension and Final}} \\
    \cmidrule(lr){2-3} \cmidrule(lr){4-5} \cmidrule(lr){6-11} \cmidrule(lr){12-14} \cmidrule(lr){15-19}
     & AS & PS & DD & PP & VQA & TC-V & TC-A & CTA & CC-V & CC & CSEC & CSBC & CSFC & VQ & DQ & SA & CSC & Final \\
    \midrule
    \multicolumn{19}{c}{\cellcolor{categorybg}\textbf{Conventional T2V Models}} \\
    Wan2.2$^\dagger$       & 87.7 & 73.5 & 57.1 & \cell{rankD}{90.7} & 40.8 & 16.1 & 35.7 & 6.3 & 31.8 & 11.3 & 25.0 & 23.9 & 21.2 & 80.6 & 73.9 & 16.0 & 23.4 & 48.5 \\
    HYVideo~1.5$^\dagger$  & 91.8 & 77.3 & 52.4 & 72.6 & 45.3 & 10.0 & 31.6 & 4.1 & 28.1 & 10.5 & 23.9 & 23.9 & 16.2 & 84.5 & 62.5 & 15.8 & 21.3 & 46.0 \\
    \midrule
    \multicolumn{19}{c}{\cellcolor{categorybg}\textbf{Native Multi-shot T2V Models}} \\
    Wan2.6                 & \cell{rankB}{96.0} & \cell{rankB}{78.5} & \cell{rankB}{68.2} & 87.5 & 81.3 & \cell{rankB}{99.5} & 50.4 & 68.1 & \cell{rankB}{99.6} & 47.4 & 91.3 & 87.4 & 66.0 & \cell{rankB}{87.2} & \cell{rankB}{77.8} & 61.7 & 81.5 & 77.1 \\
    Seedance~2.0           & 93.4 & 74.9 & 57.9 & \best{rankA}{94.9} & \cell{rankB}{86.2} & 95.3 & \cell{rankB}{76.5} & 51.8 & 96.3 & \cell{rankB}{51.0} & \cell{rankB}{96.8} & \cell{rankD}{89.5} & \best{rankA}{85.6} & 84.1 & 76.4 & \cell{rankB}{65.5} & \best{rankA}{90.6} & \cell{rankB}{79.2} \\
    Kling~3.0              & 86.9 & 71.7 & \cell{rankD}{65.5} & 89.8 & \best{rankA}{89.1} & \best{rankA}{99.6} & 50.9 & \cell{rankB}{74.5} & \best{rankA}{99.7} & \best{rankA}{56.3} & \best{rankA}{97.3} & \cell{rankB}{90.2} & \cell{rankD}{67.9} & 79.3 & \cell{rankD}{77.6} & \best{rankA}{67.7} & \cell{rankD}{85.1} & \cell{rankD}{77.4} \\
    MiniMax~H3$^\dagger$   & \cell{rankD}{94.6} & \cell{rankD}{77.5} & \best{rankA}{68.4} & \cell{rankB}{93.8} & 82.1 & 97.9 & \best{rankA}{78.8} & 42.7 & 98.4 & \cell{rankD}{49.0} & \cell{rankD}{95.9} & \best{rankA}{96.3} & \cell{rankB}{72.7} & \cell{rankD}{86.0} & \best{rankA}{81.1} & 62.7 & \cell{rankB}{88.3} & \best{rankA}{79.5} \\
    \midrule
    \multicolumn{19}{c}{\cellcolor{categorybg}\textbf{Multi-shot Specialists}} \\
    HoloCine$^\dagger$     & 77.5 & 61.3 & 57.1 & 81.6 & 80.3 & 97.1 & \cell{rankD}{52.7} & \cell{rankD}{73.2} & 97.7 & 42.8 & 89.9 & 76.4 & 55.7 & 69.4 & 69.3 & 61.8 & 74.0 & 68.6 \\
    MSM$^\dagger$          & 91.6 & 72.7 & 41.5 & 84.7 & \cell{rankD}{82.3} & \cell{rankB}{99.5} & 49.8 & \best{rankA}{87.3} & \cell{rankB}{99.6} & 42.5 & 89.9 & 73.2 & 54.1 & 82.2 & 63.1 & \cell{rankD}{65.4} & 72.4 & 70.8 \\
    VGoT$^\dagger$         & \best{rankA}{97.4} & \best{rankA}{82.0} & 44.6 & 58.6 & 69.7 & 97.5 & 46.5 & 68.1 & 98.0 & 38.2 & 35.2 & 45.7 & 49.2 & \best{rankA}{89.7} & 51.6 & 55.3 & 43.4 & 60.0 \\
 
    \bottomrule
  \end{tabular*}
\end{table*}

\subsection{Number of Shots}
\label{app:shot_count}

Table~\ref{tab:shot_count_breakdown} groups videos by the number of requested shots; \emph{Count} is the share of videos whose detected shot count equals the requested one. The two blocks differ in duration and in the systems evaluated and should be read separately. Within each block, Count falls as the number of shots grows, while Aesthetic Score stays within about two points. By system group (Figure~\ref{fig:findings}c), conventional T2V models (Kling~2.6, Hailuo~2.3, Wan2.2, HYVideo~1.5) produce the requested shot count in 32.3\%, 11.1\%, and 5.0\% of videos for two, three, and four shots, and in at most 6.7\% on the 15\,s extension. Native multi-shot T2V models decline from 81.3\% to 70.3\% and 60.0\%, mostly by adding shots: their share of videos with more shots than requested rises from 8.1\% to 24.3\%. The multi-shot specialists (HoloCine, MSM, VGoT) stay near 90\% throughout.

\begin{table}[!htbp]
  \centering
  \caption{Results by number of requested shots (0--100).}
  \label{tab:shot_count_breakdown}
  \scriptsize
  \setlength{\tabcolsep}{2pt}
  \begin{tabular*}{\linewidth}{@{\extracolsep{\fill}}cccccccccc@{}}
    \toprule
    \textbf{Shots} & \textbf{\#Videos} & \textbf{AS} & \textbf{PS} & \textbf{VQA} & \textbf{TC} & \textbf{CTA} & \textbf{CC} & \textbf{Count} & \textbf{Final} \\
    \midrule
    \multicolumn{10}{c}{\cellcolor{categorybg}\textbf{Base set, 14 systems}} \\
    2 & 2632 & 91.5 & 74.2 & 79.2 & 41.4 & 60.5 & 42.4 & 70.2 & 65.6 \\
    3 & 728 & 91.9 & 73.5 & 74.0 & 41.7 & 51.3 & 43.6 & 57.3 & 67.0 \\
    4 & 140 & 92.3 & 73.9 & 70.5 & 44.1 & 42.3 & 42.2 & 50.7 & 65.7 \\
    \midrule
    \multicolumn{10}{c}{\cellcolor{categorybg}\textbf{15\,s extension, 9 systems}} \\
    4 & 135 & 91.6 & 74.8 & 75.1 & 46.9 & 58.3 & 44.3 & 69.6 & 68.4 \\
    5 & 180 & 90.4 & 74.9 & 71.7 & 43.9 & 51.8 & 37.9 & 65.0 & 67.2 \\
    6 & 135 & 90.5 & 73.2 & 72.6 & 41.9 & 49.0 & 34.5 & 62.2 & 66.8 \\
 
    \bottomrule
  \end{tabular*}
\end{table}

\subsection{Transition Types}
\label{app:transition_breakdown}

\begin{figure*}[!t]
  \centering
  \includegraphics[width=0.95\linewidth]{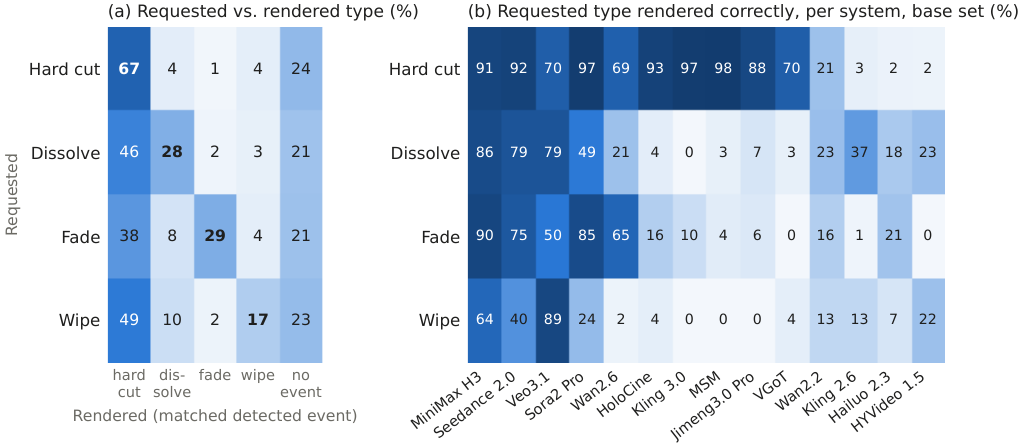}
  \caption{Transition control by type: (a) rendered type for each requested type, all videos; (b) share rendered with the requested type, per system on the base set.}
  \label{fig:transition_matrix}
\end{figure*}

Figure~\ref{fig:transition_matrix} separates Transition Control by requested type. Panel (a) pools all systems and gives the shares quoted in the main text; \emph{no event} means that no detected transition corresponds to the request. Per system (b), the systems fall into three groups: a few render gradual transitions in most cases, a larger group cuts reliably but almost always with a hard cut, and the remaining systems rarely change shots at all. Table~\ref{tab:transition_type_breakdown} lists the numbers for both parts of the benchmark. Among requested gradual transitions for which a transition event is detected, 97.4\% are hard cuts for MSM, 96.3\% for Kling~3.0, 88.5\% for HoloCine, and 85.1\% for Jimeng3.0~Pro, whereas MiniMax~H3, Seedance~2.0, and Veo3.1 substitute in fewer than 10\% of cases. On the same 250 scripts, the step from Kling~2.6 to Kling~3.0 raises shot-count accuracy by 67.2 points ($[60.8, 73.2]$) and Cut Timing Accuracy by 50.7 points, while the share of gradual transitions rendered with the requested type falls by 13.9 points ($[-19.8,-7.9]$).

\begin{table}[!htbp]
  \centering
  \caption{Share of requested transitions rendered with the requested type (\%).}
  \label{tab:transition_type_breakdown}
  \scriptsize
  \setlength{\tabcolsep}{2pt}
  \begin{tabular*}{\linewidth}{@{\extracolsep{\fill}}lcccccccc@{}}
    \toprule
     & \multicolumn{4}{c}{\textbf{Base set}} & \multicolumn{4}{c}{\textbf{15\,s ext.}} \\
    \cmidrule(lr){2-5} \cmidrule(lr){6-9}
    \textbf{Model} & Cut & Diss. & Fade & Wipe & Cut & Diss. & Fade & Wipe \\
    \midrule
    Kling~2.6        & 3.1 & 36.6 & 1.5 & 12.7 & -- & -- & -- & -- \\
    Hailuo~2.3       & 2.3 & 18.3 & 20.6 & 7.3 & -- & -- & -- & -- \\
    Wan2.2           & 21.1 & 22.5 & 16.2 & 12.7 & 9.0 & 3.3 & 0.0 & 0.0 \\
    HYVideo~1.5      & 1.6 & 22.5 & 0.0 & 21.8 & 2.0 & 10.0 & 0.0 & 2.5 \\
    Sora2~Pro        & 96.9 & 49.3 & 85.3 & 23.6 & -- & -- & -- & -- \\
    Veo3.1           & 70.3 & 78.9 & 50.0 & 89.1 & -- & -- & -- & -- \\
    Wan2.6           & 68.8 & 21.1 & 64.7 & 1.8 & 100.0 & 0.0 & 0.0 & 0.0 \\
    Jimeng3.0~Pro    & 88.3 & 7.0 & 5.9 & 0.0 & -- & -- & -- & -- \\
    Seedance~2.0     & 92.2 & 78.9 & 75.0 & 40.0 & 90.0 & 73.3 & 86.7 & 17.5 \\
    Kling~3.0        & 96.9 & 0.0 & 10.3 & 0.0 & 99.0 & 3.3 & 3.3 & 2.5 \\
    MiniMax~H3       & 90.6 & 85.9 & 89.7 & 63.6 & 80.0 & 86.7 & 86.7 & 47.5 \\
    HoloCine         & 93.0 & 4.2 & 16.2 & 3.6 & 98.0 & 0.0 & 10.0 & 2.5 \\
    MSM              & 97.7 & 2.8 & 4.4 & 0.0 & 98.0 & 0.0 & 0.0 & 2.5 \\
    VGoT             & 70.3 & 2.8 & 0.0 & 3.6 & 88.0 & 3.3 & 0.0 & 7.5 \\
    \midrule
    All systems      & 63.8 & 30.8 & 31.4 & 20.0 & 73.8 & 20.0 & 20.7 & 9.2 \\
 
    \bottomrule
  \end{tabular*}
\end{table}

\subsection{Scene Change and Cut Rendering}
\label{app:scene_cut}
Table~\ref{tab:scene_cut} gives the numbers behind the scene-change finding. For each system, we compute the share of requested cuts that are rendered in single-scene and in multi-scene base-set scripts, and the difference after stratifying by duration, shot count, and requested transition type. In the simplest cell, a 4\,s script with two shots joined by a hard cut, Kling~2.6 produces two shots in 15 of 22 multi-scene scripts and in none of 37 single-scene scripts, and Wan2.6 in 86.4\% against 27.0\%. The gap is not a detection artifact: on the human-labeled diagnostic set (Appendix~\ref{app:transition_validation}), the detector finds a cut in 93.3\% of single-scene and 97.4\% of multi-scene videos that contain one, and reports no cut in any video without one.

\begin{table}[!htbp]
  \centering
  \caption{Share of requested cuts rendered in single- and multi-scene base-set scripts (\%). Diff.\ is the single- minus multi-scene difference after stratifying by duration, shot count, and requested transition type, so it can differ from the difference of the two pooled columns.}
  \label{tab:scene_cut}
  \footnotesize
  \setlength{\tabcolsep}{4pt}
  \begin{tabular*}{\linewidth}{@{\extracolsep{\fill}}lcccc@{}}
    \toprule
    \textbf{Model} & \textbf{Single} & \textbf{Multi} & \textbf{Diff.} & \textbf{95\% CI} \\
    \midrule
    Wan2.2 & 25.7 & 70.1 & $-47.4$ & [$-57.3$, $-36.5$] \\
    Kling~2.6 & 11.5 & 52.3 & $-45.7$ & [$-56.0$, $-36.3$] \\
    HYVideo~1.5 & 14.9 & 50.0 & $-40.5$ & [$-50.8$, $-32.1$] \\
    Hailuo~2.3 & 6.8 & 29.3 & $-24.6$ & [$-33.7$, $-16.7$] \\
    Wan2.6 & 76.4 & 97.1 & $-17.2$ & [$-24.9$, $-10.1$] \\
    Veo3.1 & 79.7 & 94.3 & $-13.2$ & [$-20.8$, $-6.6$] \\
    Jimeng3.0~Pro & 76.4 & 82.2 & $-6.0$ & [$-14.7$, $3.1$] \\
    MiniMax~H3 & 92.6 & 97.7 & $-5.3$ & [$-11.5$, $0.4$] \\
    HoloCine & 90.5 & 97.1 & $-4.9$ & [$-10.1$, $0.7$] \\
    Seedance~2.0 & 95.3 & 98.3 & $-3.2$ & [$-7.5$, $1.7$] \\
    Sora2~Pro & 97.3 & 98.9 & $-2.0$ & [$-5.2$, $0.7$] \\
    MSM & 99.3 & 99.4 & $0.0$ & [$-1.6$, $1.5$] \\
    Kling~3.0 & 97.3 & 98.3 & $+0.2$ & [$-2.9$, $3.6$] \\
    VGoT & 98.6 & 96.0 & $+1.2$ & [$-2.4$, $4.6$] \\
    \bottomrule
  \end{tabular*}
\end{table}

\ifdefined\CameraFigInMain\else
\begin{figure*}[!t]
  \centering
  \includegraphics[width=0.92\linewidth]{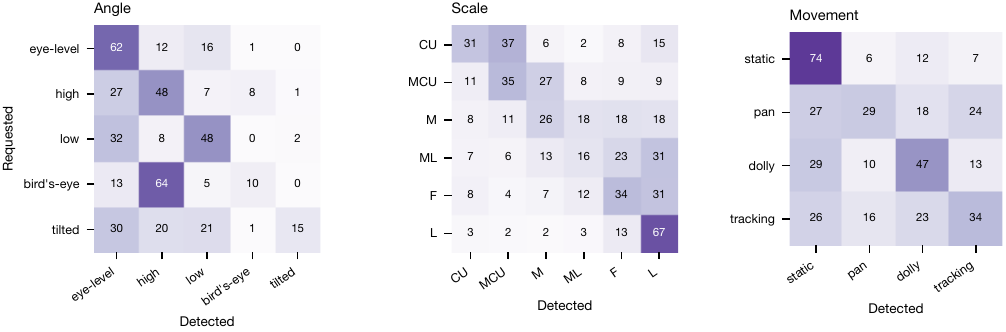}
  \caption{Requested against detected camera state on rendered shots (row-normalized, \%).}
  \label{fig:camera_confusion}
\end{figure*}
\fi

\subsection{Scene-Change Intervention}
\label{app:c1_intervention}
We select 24 base-set scripts that are single-scene, 4\,s long, and contain two shots joined by a hard cut, with a second shot centered on an entity rather than on the location. For each script, version B adds a second location to the global context and moves the second shot there, rewriting only the location phrases; entities, actions, camera states, timestamps, and the requested transition are the same as in the original version A. Wan2.6, Kling~2.6, and Wan2.2 each generate both versions once with a fixed seed, and the first 4\,s of every clip are labeled manually, blind to the system, the version, and the script. In Table~\ref{tab:c1_intervention}, a video has a \emph{boundary} when it contains at least two shots, +/$-$ counts the pairs where only B / only A has one, and $p$ is a two-sided sign test over these pairs. Most added boundaries are not the requested hard cut: B alone has a hard cut in 13 pairs, against 32 for a boundary of any type, and Kling~2.6 renders 11 of its 15 version-B boundaries as dissolves.
Because the three systems share the same scripts, we also test at the script level, summing the boundary changes of the three systems per script: boundaries increase in 17 scripts, decrease in one, and are unchanged in six (exact sign-flip test, $p=5\times10^{-5}$). Pooled over systems, the boundary rate rises by 40.3 points (95\% CI [26.4, 52.8]), and counting boundaries with the evaluation pipeline instead of the manual labels gives the same conclusion (Appendix~\ref{app:transition_validation}).

\begin{table}[!htbp]
  \centering
  \caption{Scene-change intervention on 24 single-scene scripts per system.}
  \label{tab:c1_intervention}
  \footnotesize
  \setlength{\tabcolsep}{4pt}
  \begin{tabular*}{\linewidth}{@{\extracolsep{\fill}}lcccccc@{}}
    \toprule
    System & \multicolumn{2}{c}{Boundary} & +/$-$ & $p$ & \multicolumn{2}{c}{Hard cut} \\
    & A & B & & & A & B \\
    \midrule
    Wan2.6 & 13 & 18 & 8/3 & 0.23 & 11 & 16 \\
    Kling~2.6 & 1 & 15 & 14/0 & $1.2\times10^{-4}$ & 0 & 1 \\
    Wan2.2 & 7 & 17 & 10/0 & 0.002 & 6 & 8 \\
    \midrule
    Pooled & 21 & 50 & 32/3 & $4\times10^{-7}$ & 17 & 25 \\
    \bottomrule
  \end{tabular*}
\end{table}

\subsection{Camera States}
\label{app:camera_breakdown}

Table~\ref{tab:camera_state_breakdown} reports Camera Control by requested state, pooled over systems. The \emph{Score} column covers the requested shots that have a detected counterpart, which separates camera execution from shot structure; the last column counts missing shots as zero, as the metric does. Tilt and handheld have no direct FilmOps class and earn at most partial credit (Appendix~\ref{app:semantic_metrics}), so we draw no conclusions from them; skipping the movement aspect on these 184 shots raises system-level CC by 1.7 points on average and leaves the ranking nearly unchanged (SRCC 0.98).
Figure~\ref{fig:camera_confusion} shows where the errors go. A static camera is requested in 21.2\% of the rendered shots and detected in 39.7\%, and pan, dolly, and tracking requests are detected as static in 25--29\% of cases. Framing drifts toward wider shots: long shots are requested in 15.5\% and detected in 26.8\%. For camera angle, eye-level is the most common wrong detection (32.9\% of angle errors), and a requested bird's-eye view is detected as an ordinary high angle in 64\% of shots.

\begin{table}[!htbp]
  \centering
  \caption{Camera Control by requested camera state (0--100).}
  \label{tab:camera_state_breakdown}
  \footnotesize
  \setlength{\tabcolsep}{4pt}
  \begin{tabular*}{\linewidth}{@{\extracolsep{\fill}}llccc@{}}
    \toprule
    \textbf{Aspect} & \textbf{Requested state} & \textbf{\#Shots} & \textbf{Score} & \textbf{Incl.\ missing} \\
    \midrule
    Angle    & eye-level & 3668 & 76.0 & 66.0 \\
             & high-angle & 2395 & 70.3 & 60.3 \\
             & low-angle & 2143 & 64.5 & 53.9 \\
             & bird's-eye-view & 698 & 44.7 & 39.1 \\
             & tilted & 1354 & 14.3 & 12.0 \\
    \midrule
    Scale    & long & 1592 & 66.3 & 56.7 \\
             & medium-close-up & 1620 & 53.4 & 44.9 \\
             & full & 1194 & 52.6 & 46.1 \\
             & close-up & 1477 & 49.0 & 39.9 \\
             & medium & 2379 & 39.8 & 35.1 \\
             & medium-long & 1253 & 34.6 & 28.8 \\
             & extreme-close-up & 743 & 26.2 & 23.4 \\
    \midrule
    Movement & static & 2183 & 73.6 & 62.7 \\
             & dolly & 1281 & 46.8 & 39.6 \\
             & pan & 1956 & 31.5 & 26.8 \\
             & zoom & 1236 & 30.1 & 25.8 \\
             & tracking & 1326 & 26.6 & 23.2 \\
             & tilt & 1138 & 19.7 & 16.6 \\
             & handheld & 1138 & 2.7 & 2.4 \\
 
    \bottomrule
  \end{tabular*}
\end{table}

\subsection{Validity and Conditional Score}
\label{app:validity_split}
Each structure-dependent sub-metric is the product of validity, the share of applicable videos that can be evaluated, and the score on valid videos. On face consistency, VGoT, Veo3.1, Kling~3.0, and Jimeng3.0~Pro are within 1.4 points of each other in the product, although their validity ranges from 72.8\% to 96.3\%. Among the 91 pairs of systems, the two factors differ significantly in opposite directions while the products do not in 9 pairs for Transition Control, 6 for face, 4 for background, and 3 for entity consistency.

\subsection{Difficulty Types and Topics}
\label{app:slices}

Table~\ref{tab:per_difficulty} reports the base-set results by difficulty type and by topic, averaged over the 14 systems. Visual and Dynamic Quality vary little across difficulty types. Multi-scene scripts obtain higher Semantic Alignment and Cross-Shot Consistency than single-scene ones: a change of location gives the model a strong cue to start a new shot, whereas consecutive shots of the same scene are more often merged into one, which invalidates the consistency metrics. Differences across topics are smaller than differences across systems; Sports and Cosmic Space are the hardest topics.

\begin{table}[!htbp]
  \centering
  \caption{Base-set results by difficulty type and topic (0--100), averaged over systems.}
  \label{tab:per_difficulty}
  \footnotesize
  \setlength{\tabcolsep}{3pt}
  \begin{tabular*}{\linewidth}{@{\extracolsep{\fill}}lcccccc@{}}
    \toprule
     & \textbf{\#Scripts} & \textbf{VQ} & \textbf{DQ} & \textbf{SA} & \textbf{CSC} & \textbf{Final} \\
    \midrule
    \multicolumn{7}{c}{\cellcolor{categorybg}\textbf{Difficulty type}} \\
    Cross-entity, multi-scene & 64 & 83.0 & 72.8 & 56.7 & 61.9 & 68.6 \\
    Cross-entity, single scene & 67 & 83.3 & 72.2 & 52.9 & 49.4 & 64.4 \\
    No cross-entity, multi-scene & 60 & 82.6 & 72.0 & 57.4 & 76.5 & 72.1 \\
    No cross-entity, single scene & 59 & 82.3 & 74.4 & 52.9 & 52.4 & 65.5 \\
 
    \midrule
    \multicolumn{7}{c}{\cellcolor{categorybg}\textbf{Topic}} \\
    Cultural Customs & 16 & 85.3 & 72.6 & 58.6 & 61.4 & 69.5 \\
    Leisure Hobbies & 16 & 86.8 & 71.9 & 56.1 & 59.1 & 68.5 \\
    Social Interaction & 19 & 85.4 & 72.8 & 59.3 & 56.1 & 68.4 \\
    Transportation & 13 & 84.5 & 75.6 & 56.6 & 52.4 & 67.3 \\
    Natural Phenomena & 14 & 80.9 & 72.2 & 53.7 & 60.6 & 66.9 \\
    Wildlife & 18 & 84.6 & 69.7 & 54.4 & 58.8 & 66.8 \\
    Daily Living & 20 & 83.7 & 73.4 & 58.0 & 52.1 & 66.8 \\
    Artistic Styles & 16 & 80.0 & 73.3 & 52.9 & 60.5 & 66.7 \\
    Urban Scenes & 20 & 83.4 & 74.7 & 56.3 & 51.2 & 66.4 \\
    Fantasy & 16 & 82.6 & 73.8 & 54.8 & 50.4 & 65.4 \\
    Technology & 13 & 81.6 & 76.7 & 52.4 & 50.3 & 65.2 \\
    Cinematic & 23 & 79.4 & 71.9 & 52.8 & 55.1 & 64.8 \\
    Landscapes & 17 & 84.3 & 71.3 & 53.7 & 44.5 & 63.4 \\
    Sports & 20 & 80.0 & 70.2 & 52.6 & 47.4 & 62.6 \\
    Cosmic Space & 9 & 79.5 & 75.9 & 49.2 & 44.1 & 62.2 \\
 
    \bottomrule
  \end{tabular*}
\end{table}

\end{document}